\documentclass[11pt]{article}
\usepackage{amssymb}
\usepackage{amsthm}
\usepackage{amsmath}
\usepackage{float,booktabs}
\usepackage{float} 
\usepackage{graphicx}
\usepackage{natbib}
\usepackage{url}
\usepackage{xcolor}
\usepackage{soul,color,colortbl}
\usepackage{array,multirow}
\usepackage{cancel}
\usepackage[top=0.9in, bottom=1.2in, left=0.875in, right=0.875in]{geometry}
\usepackage[linesnumbered,ruled]{algorithm2e}
\usepackage{tikz}
\usetikzlibrary{er,positioning,arrows.meta,calc}
\usepackage[labelformat=simple]{subcaption}
\usepackage{bm}
\usetikzlibrary{shapes,arrows,shadows}
\usepackage{bbm}
\usepackage{hyperref}
\usepackage{rotating}
\usepackage{tablefootnote}
\usepackage{xurl} 
\usepackage[titletoc,toc,title]{appendix}
\usepackage{nccmath}

\title{Efficient and Interpretable Transformer for Counterfactual Fairness}
\author{Panyi Dong \thanks{Acturial and Risk Mangament Sciences, University of Illinois Urbana-Champaign, Champaign, Illinois, USA. Email: \texttt{panyid2@illinois.edu}}  \and Zhiyu Quan
\thanks{Acturial and Risk Mangament Sciences, University of Illinois Urbana-Champaign, Champaign, Illinois, USA. Email: \texttt{zquan@illinois.edu}}}

\begin{document}

\maketitle

\begin{abstract}
The growing reliance of machine learning models in high-stakes, highly regulated domains such as finance and insurance has created a growing tension between predictive performance, interpretability, and regulatory fairness requirements. In these settings, models are expected not only to deliver reliable predictions but also to provide transparent decision rationales and comply with strict fairness requirements. Existing fairness-aware learning methods for tabular data, however, often focus primarily on group-level fairness metrics or depend on explicit and structural causal model assumptions that are challenging to validate in practice. Meanwhile, attention-based transformers offer powerful mechanisms for modeling complex data relationships as demonstrated in various language tasks, yet their attention mechanisms alone do not ensure counterfactually fair predictions, even when combined with fairness-aware techniques. To address these limitations, we propose the Feature Correlation Transformer (FCorrTransformer), an attention-light architecture tailored for tabular data. In this design, the attention matrix admits a direct statistical interpretation as pairwise feature dependencies, enhancing both interpretability and efficiency. Leveraging this structure, we introduce Counterfactual Attention Regularization (CAR), a framework that enforces group-invariant fair representations of sensitive features at the attention level, promoting counterfactually fair predictions without relying on explicit causal assumptions. Empirical evaluations on imbalanced classification and regression benchmarks demonstrate that FCorrTransformer combined with CAR achieves strong counterfactual fairness while maintaining competitive predictive performance and substantially reducing model complexity compared with standard transformer-based baselines. Overall, this work bridges a critical gap between fairness theory and machine learning models, offering a practical framework for responsible AI in regulatory-sensitive domains.
\end{abstract}

\newpage

\section{Introduction}\label{sec:intro}

The rapid adoption of machine learning (ML) has transformed business operations across many industries, enabling organizations to use data to automate and improve complex processes. At the same time, ML has also raised serious concerns about governance, transparency, and ethics, particularly in heavily regulated industries such as finance and insurance. In these domains, automated operations can have significant financial and social consequences for individuals, making issues of discrimination and accountability especially critical. Regulators have responded to these concerns through the introduction of legal and regulatory frameworks governing the use of ML in business operations. For example, regulatory initiatives such as \textit{Protecting Consumers from Unfair Discrimination in Insurance Practices} (SB21-169)\footnote{\url{https://doi.colorado.gov/for-consumers/sb21-169-protecting-consumers-from-unfair-discrimination-in-insurance-practices}} proposed by the State of Colorado, the \textit{NAIC Model Bulletin}\footnote{\url{https://content.naic.org/sites/default/files/2023-12-4\%252520Model\%252520Bulletin\_Adopted\_0.pdf}} issued by the National Association of Insurance Commissioners (NAIC), and \textit{Insurance Circular Letter No. 7}\footnote{\url{https://www.dfs.ny.gov/industry-guidance/circular-letters/cl2024-07}} from the New York State Department of Financial Services all prohibit the use of ``unfair discrimination'' in insurance decision-making. Although formal definitions of algorithmic fairness/bias are still evolving, the insurance industry already operates under strict regulatory constraints, where even subtle forms of bias in ML may lead to regulatory penalties. In addition to fairness, interpretability is a fundamental requirement for the adoption of ML in insurance. Insurers must be able to explain how and why certain decisions are made, both for internal business purposes and to meet regulatory expectations. This requirement is particularly important given the societal role of insurance, which directly affects access to essential services such as healthcare, housing, and financial security. The use of sensitive personal information, such as age, gender, or socioeconomic characteristics, further intensifies concerns about affordability and unequal treatment, especially for minority and vulnerable populations. As a result, there is a strong demand for interpretable ML models with rigorous fairness guarantees in financial domains such as insurance.

To achieve algorithmic fairness, a wide range of bias mitigation strategies, from simple preprocessing methods to more complex in-processing approaches, have been proposed for tabular learning \citep{caldersThreeNaiveBayes2010, kamishimaFairnessAwareClassifierPrejudice2012, zafarFairnessConstraintsMechanisms2017}. In recent years, a variety of more sophisticated mitigation frameworks have been proposed, including adversarial learning approaches \citep{zhangMitigatingUnwantedBiases2018}, fairness-aware loss functions \citep{qianReducingGenderBias2019}, and the construction of embeddings invariant to sensitive demographic categories \citep{luGenderBiasNeural2020}. In this work, we focus on counterfactual fairness \citep{kusnerCounterfactualFairness2017}, which requires that insurers do not discriminate against individuals based on sensitive features or their membership in protected demographic categories. Under this notion, two policyholders who are identical in all respects except for their sensitive features should receive identical treatment or model predictions. This provides a practical and actionable definition of fairness at the individual level. Beyond its conceptual alignment with our proposed framework, counterfactual fairness offers a more operational perspective compared to group-based fairness criteria. Group fairness metrics are often interpreted as requiring compensatory adjustments across demographic categories, which may conflict with the actuarial principle of risk-based pricing and can inadvertently introduce new forms of unfairness or affordability concerns in insurance practice. Importantly, our framework does not rely on the exclusion of sensitive features. While regulatory efforts such as SB21-169 restrict the use of features like credit scores, these features may still carry predictive information relevant to underlying risk. Simply removing them does not eliminate bias; rather, it can lead to the emergence of proxy features. For instance, features such as mortgage payment history, credit limits, and credit age can collectively reconstruct information similar to a credit score. As a result, bias may persist in indirect forms. Instead, our approach explicitly incorporates sensitive features within a principled framework that mitigates their unfair influence while preserving predictive performance. By addressing bias at a structural level rather than through superficial feature removal, the proposed framework provides a more robust and effective solution to fairness in actuarial modeling.

Bias mitigation can be conceptually understood as modifying feature dependencies by suppressing or reshaping dependencies of sensitive features in accordance with societal and regulatory objectives. One of the most effective architectures for capturing such dependencies is the attention mechanism introduced in transformers \citep{vaswaniAttentionAllYou2017}, which has driven the rapid development of Large Language Models (LLMs), achieving remarkable success in a wide range of language tasks due to its ability to capture complex semantic dependencies. Motivated by this success, a growing body of work has extended transformer-based architectures to tabular data modeling \citep{gorishniyRevisitingDeepLearning2021, zhuXTabCrosstablePretraining2023, hollmannAccuratePredictionsSmall2025}. The attention mechanism provides a natural and flexible framework for representing feature dependencies, making it an appealing tool for identifying and mitigating unintended biased dependencies. While attention-based debiasing methods have been explored in language \citep{gaciDebiasingPretrainedText2022, haqueFinetuningLLMsCrossAttentionbased2025} and computer vision \citep{qiangFairnessAwareVisionTransformer2025} tasks, relatively little work has addressed this problem in tabular settings. Moreover, existing approaches to counterfactual fairness largely fall into two categories. The first relies on explicit causal models to characterize feature dependencies \citep{kusnerCounterfactualFairness2017, raoCounterfactualAttentionLearning2021, robertsonFairPFNTabularFoundation2025}, which often require strong structural assumptions that may be difficult to validate in practice. The second category applies post hoc adjustments to learned representations or attention outputs \citep{raoCounterfactualAttentionLearning2021, duFairnessRepresentationNeutralization2021, wangRobustNaturalLanguage2023}. However, enforcing fairness constraints at the attention level alone does not generally guarantee counterfactually fair predictions under standard transformer architectures. These limitations highlight the need for assumption-light, attention-based bias mitigation frameworks that are both interpretable and capable of achieving counterfactual fairness in tabular data settings.

We reformulate fair prediction as a problem of fair attention, which aligns with the notion of counterfactual fairness and enables more intuitive interpretation and regularization. To address the limitation that fair attention may still lead to discriminatory predictions under classical attention mechanisms, we propose a modified transformer architecture, termed \textit{Feature Correlation Transformer} (FCorrTransformer), whose attention matrix offers a more direct interpretation as feature dependencies. Building on this architecture, we introduce \textit{Counterfactual Attention Regularization} (CAR), a method that enforces fairness directly at the level of attention, without relying on explicit causal models. The proposed framework applies to both classification and regression tasks. We further validate the proposed framework on both synthetic and real-world datasets, evaluating its performance, fairness, interpretability, and computational efficiency.

The remainder of the paper is structured as follows: Section \ref{sec:lit} reviews the related literature. Section \ref{sec:method} introduces the transformer architecture and fairness regularization methods we proposed. Section \ref{sec:exp} demonstrates the empirical results and the interpretation of the proposed frameworks. Section \ref{sec:con} concludes.  

\section{Related Work}\label{sec:lit}

\textbf{Attention-based Tabular Learning:} 
TabTransformer \citep{huangTabTransformerTabularData2020} represents categorical features' levels as tokens and applies self-attention to model relationships among categories, while treating numerical features outside the attention mechanism. FT-Transformer \citep{gorishniyRevisitingDeepLearning2021} extends this idea by embedding numerical features into high-dimensional representations, enabling joint attention over both categorical and continuous features. SAINT \citep{somepalliSAINTImprovedNeural2021} further augments standard attention blocks with an inter-sample attention (MISA) module, capturing dependencies not only across features but also across samples within a batch. TabNet \citep{arikTabNetAttentiveInterpretable2021} adopts a sequential attention mechanism to iteratively perform feature selection and transformation, offering improved interpretability of the learning process. XTab \citep{zhuXTabCrosstablePretraining2023} leverages a shared attention backbone to learn cross-table feature dependencies across multiple tabular datasets and tasks. TabPFN \citep{hollmannAccuratePredictionsSmall2025} reformulates tabular learning as an in-context learning problem, training a transformer-based tabular foundation model on large-scale synthetic datasets generated from causal models to approximate complex tabular distributions. Alternatively, rather than designing tabular-specific architectures, TabLLM \citep{hegselmannTabLLMFewshotClassification2023} and TP-BERTa \citep{yanMakingPretrainedLanguage2024} embed tabular data into natural language representations, allowing pretrained language models to process tabular numerical inputs directly. 

Across existing tabular transformer architectures, raw tabular features are first projected into high-dimensional embedding vectors before being processed by attention mechanisms. However, this process may impose a structural mismatch on the data. Unlike text, where tokens have inherent semantic relationships, tabular data is typically a collection of independent and heterogeneous features that lack a natural sequential or spatial order. In addition, when a simple numerical value, such as a dollar amount or an age, is projected into high-dimensional embedding vectors, the model effectively treats a basic magnitude as a complex semantic concept, forcing the Transformer to laboriously re-learn ordinal relationships that the original data carries instinctively. This transformation often results in a massive explosion of parameters, which can lead to rapid overfitting on smaller tabular datasets where the model has enough capacity to memorize noise rather than identify generalizable patterns. 

\textbf{Attention-based Bias Mitigation:} 
Bias mitigation has become a critical component for the responsible deployment of ML systems in real-world applications. 
A variety of mitigation strategies have been proposed, including adversarial learning approaches \citep{zhangMitigatingUnwantedBiases2018}, fairness-aware loss functions \citep{qianReducingGenderBias2019}, and the construction of embeddings invariant to sensitive demographic categories \citep{luGenderBiasNeural2020}.
The widespread adoption of transformer architectures has motivated the development of bias mitigation techniques in language and computer vision domains that leverage attention mechanisms.
\citet{attanasioEntropybasedAttentionRegularization2022} propose Entropy-based Attention Regularization (EAR), which maximizes entropies of attention scores, thereby reducing algorithmic bias induced by low contextualization in language tasks.
\citet{mehrabiAttributingFairDecisions2022} propose a post-hoc framework that identifies features contributing to discriminatory outcomes and mitigates the bias by artificially assigning zero to the corresponding attention weights.
Attention-Debiasing (AttenD) \citep{gaciDebiasingPretrainedText2022} introduces an additional regularization term in the loss function that penalizes disparities in attention weights from non-sensitive words toward sensitive demographic words. 
REsidual Attention Debiasing (READ) \citep{wangRobustNaturalLanguage2023} adopt a product-of-experts (PoE) framework consisting of main and bias-specific attention modules. In this framework, biased information is isolated within the bias attention, allowing the main model to generate fair predictions in language tasks.
Cross-Attention-based Weight Decay (CrAWD) \citep{haqueFinetuningLLMsCrossAttentionbased2025} adjusts the cross-attention weights between the input sequence and a reference sensitive sequence by a pre-defined decay factor during model training. \citet{qiangFairnessAwareVisionTransformer2025} propose the Debiased Self-Attention (DSA) framework, a two-stage adversarial procedure for mitigating social bias in Vision Transformers (ViTs). In DSA, a bias-only model is first trained to explicitly identify the influence of sensitive features, after which a debiased model is trained to reduce or remove this influence.
Across the proposed methods, imposing artificial decay on attention weights \citep{mehrabiAttributingFairDecisions2022, haqueFinetuningLLMsCrossAttentionbased2025} may disrupt the training process and still leave residual biased information in the learned representations. In contrast, regularizing the loss function at model predictions \citep{qianReducingGenderBias2019, wangRobustNaturalLanguage2023} constrains the flexibility of the entire model, which may adversely affect predictive performance.

\section{Methodology}\label{sec:method}

To enable counterfactual fairness learning via attention mechanisms, we propose a framework integrating FCorrTransformer, optimized for capturing complex feature dependencies in tabular data, with a CAR scheme designed to learn counterfactually fair representations.

The problem of fairness learning can be formulated as: Let $\mathcal{D}=(\bm{X}, \bm{y})=\{\bm{X}_{n}, y_{n}\}_{n=1}^{N}$ denote a dataset consisting of $N$ independent observations. For each observation $n$, the data consists of two components: feature vectors $\bm{X}_{n} \in\mathcal{X}$ with $p$ features and response feature $y_{n}\in\mathcal{Y}$. 
Regarding the type of features, the feature matrix $\bm{X}$ can be decomposed into categorical features $\bm{X}^{Cat}\in\mathcal{X}^{Cat}$ with $p^{Cat}$ features and continuous features $\bm{X}^{Con}\in\mathcal{X}^{Con}$ with $p^{Con}$ features ($p=p^{Cat}+p^{Con}$).
In the context of fair learning, the feature matrix $\bm{X}$ can be also decomposed into sensitive features $\bm{X}_{s}\in\mathcal{X}_s$ and non-sensitive features $\bm{X}_{r}\in\mathcal{X}_{r}$, which may be correlated with sensitive features $\bm{X}_{s}$. In this work, we assume that the sensitive features are categorical, either binary or multi-class. In our setting, the continuous sensitive features can be grouped into categories. Furthermore, for the sake of simplicity, we assume there is only one sensitive feature, which can be extended to multi-sensitive features scenarios if needed. Hence, for $n$-th observation, the sensitive feature $x_{n, s}\in\mathcal{S}=\{s_i\}_{i=1}^{C^s}$ where $s_i$ denotes the $i$-th unique category of the sensitive feature, and $C^s$ is number of unique categories of the sensitive feature. The supervised ML model $f:\bm{X}\rightarrow\bm{y}$ maps the combined features to the response feature, producing predictions $\hat{\bm{y}}=f(\bm{X})$. Under our formulation, a ML model $f$ is considered counterfactually fair if $p(\hat{y}|\bm{X}_r=\bm{x}_r, X_s=s_i)=p(\hat{y}|\bm{X}_r=\bm{x}_r, X_s=s_j)$ for any input $\bm{x}_r$ and sensitive categories $s_i, s_j\in\mathcal{S}$ \citep{kusnerCounterfactualFairness2017}. In the context of insurance pricing, counterfactual fairness can be interpreted as: a policyholder should be priced identically to another policyholder (which may not exist in real life or in observed data) with identical features except for the sensitive feature. Please refer to Appendix \ref{appendix_sec:notation} for the notations used in the paper.

\subsection{\texttt{FCorrTransformer}: Feature Correlation Transformer}\label{subsec:FCorr}

We design FCorrTransformer with the following consideration in mind that: 1) Since we aim to equate fair predictions with fair attention, classical transformer architecture must be modified to prevent biased information from bypassing the attention mechanism. 2) Unlike classical tabular transformers that utilize high-dimensional embeddings \citep{gorishniyRevisitingDeepLearning2021, somepalliSAINTImprovedNeural2021, zhuXTabCrosstablePretraining2023}, we argue that raw feature values in tabular data already contain sufficient information for effective attention. Therefore, the attention mechanism in FCorrTransformer is designed to efficiently learn representations that remain close to the original inputs while explicitly capturing feature dependencies.

\begin{figure}[!ht]
\centering
\begin{subfigure}[t]{0.3\textwidth}
    \vspace{20pt}
    \centering
    \includegraphics[width=\textwidth]{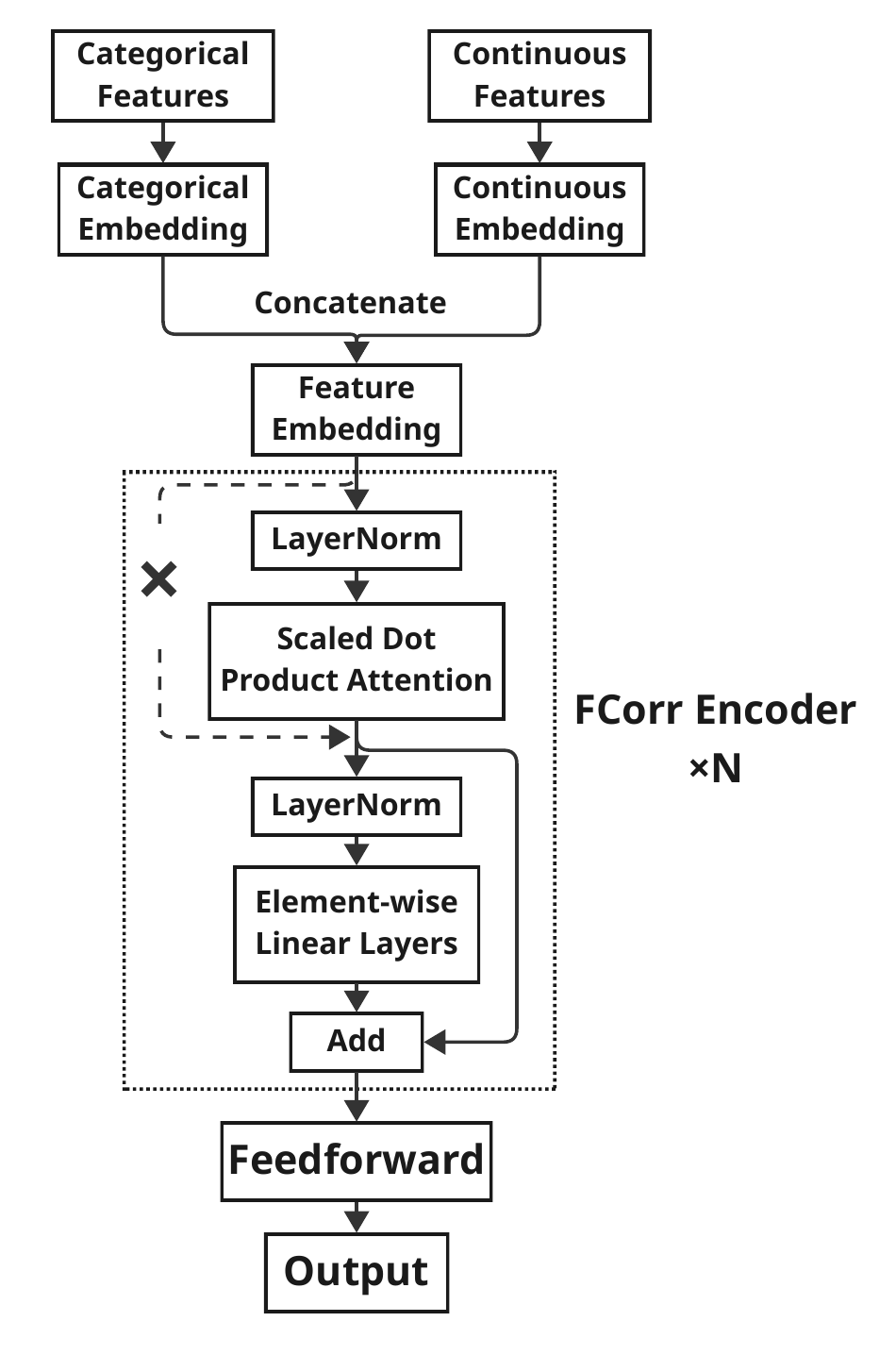}
    \caption{Overall architecture}
    \label{fig:FCorr_architecture}
\end{subfigure}
\begin{subfigure}[t]{0.5\textwidth}
    \vspace{0pt}
    \centering
    \includegraphics[width=\textwidth]{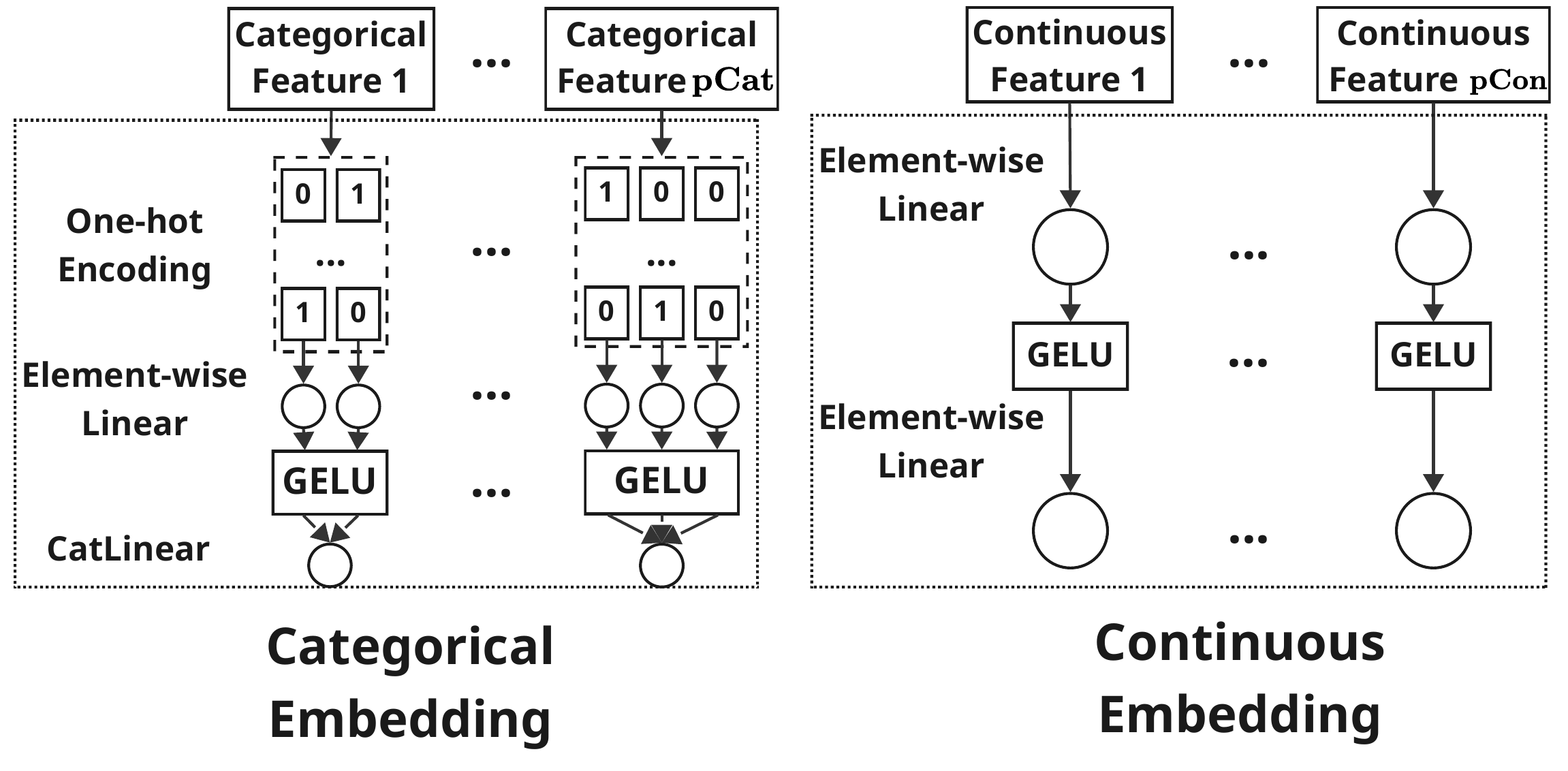}
    \caption{Embedding architecture}
    \label{fig:FCorr_embedding}
    \vspace{0.5em}
    \includegraphics[width=0.6\textwidth]{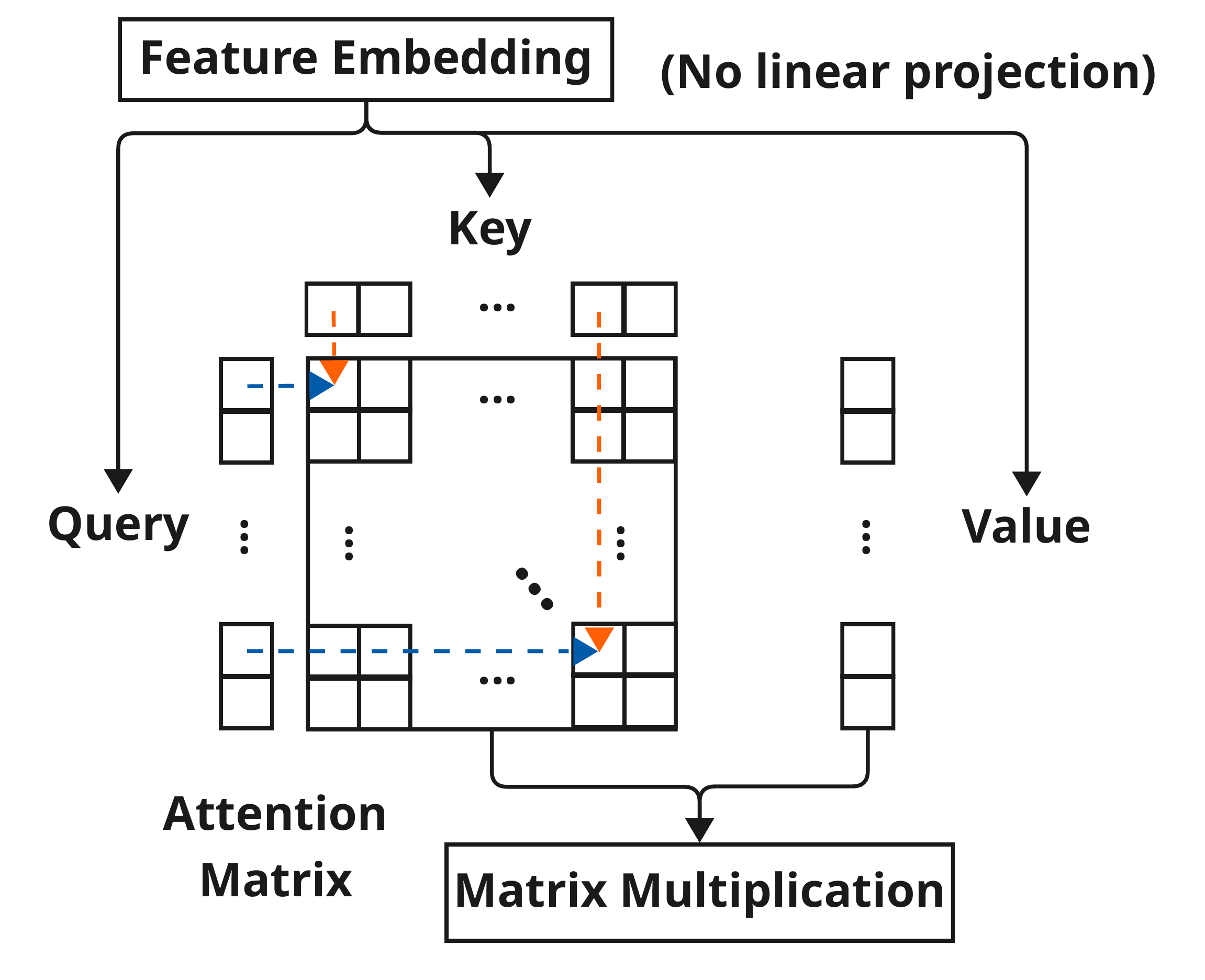}
    \caption{Attention architecture}
    \label{fig:FCorr_attention}
\end{subfigure}
\caption{Architecture Design of FCorrTransformer}
\label{fig:FCorrTransformer}
\end{figure}

\textbf{Overall Architecture:} As illustrated in Figure \ref{fig:FCorr_architecture}, the overall FCorrTransformer architecture follows an encoder-only structure similar to FT-Transformer. Both numerical and categorical features are first passed through corresponding one-dimensional embedding layers and then concatenated to form the feature embedding. In each FCorr encoder layer, layer normalization (LayerNorm) is applied before the scaled dot-product attention and the linear transformations (Element-Wise Linear layers), consistent with FT-Transformer, which has been shown to provide superior performance in our prior empirical analysis. Since the feature representations are one-dimensional, we adopt Element-Wise Linear layers instead of fully connected layers to avoid introducing artificial feature interactions through the representation space. Furthermore, to prevent biased information from bypassing the attention mechanism, we remove the residual connections around the attention block while retaining residual connections in the Element-Wise Linear layers. Finally, a standard multi-layer feedforward network is appended after the encoder stack to produce the final predictions.

\textbf{Embedding Architecture:} Figure \ref{fig:FCorr_embedding} illustrates the embedding design in FCorrTransformer. 
For categorical features, each feature is one-hot encoded to vectors of unique categories without dropping the base category and then passed through an Element-Wise Linear layer followed by a Gaussian Error Linear Unit (GELU) activation \citep{hendrycks2016gaussian}. We utilize the GELU activation function, as it provides smooth gradient behavior around zero, is widely adopted in modern transformer architectures, and demonstrated the best performance in our preliminary experiments. To ensure dimensional consistency with the original inputs, the projected activations for each categorical feature are subsequently aggregated through a linear layer, termed \textit{CatLinear} layer, producing a single activation per categorical feature.
For the $n$-th observation, the $i$-th categorical feature $x^{Cat}_{n,i}$ ($i=1,2,...,p^{Cat}$) with $C^{Cat}_i$ unique categories is one-hot encoded, following $OH_i(x^{Cat}_{n, i})=\bm{v}_{n,i}$ where $\bm{v}_{n,i}\in\{0, 1\}^{C^{Cat}_i\times 1}$ is a unit vector whose element $v^{j}_{n,i}$ is the indicator function of $x^{Cat}_{n,i}$ equals $j$th categories. The one-hot encoded categorical features for the $n$-th observation thus can be expressed as 
\begin{align*}
OH(\bm{x}^{Cat}_{n})&=[OH_1(x^{Cat}_{n, 1}), OH_2(x^{Cat}_{n, 2}), ..., OH_{p^{Cat}}(x^{Cat}_{n, p^{Cat}})]\\
&=[\bm{v}_{n,1}, \bm{v}_{n,2}, ..., \bm{v}_{n,p^{Cat}}]=\bm{v}_n\in\{0, 1\}^{p_{OH}^{Cat}\times1}
\end{align*}
where $p_{OH}^{Cat}=\sum_{i=1}^{p^{Cat}}C^{Cat}_{i}$ denotes total unique categories. The one-hot encoded features are then passed through a Element-Wise Linear layer $ELinear^{Cat}$ with weights $\bm{W}^{Cat}_1\in\mathbb{R}^{p_{OH}^{Cat}\times1}$ and biases $\bm{b}^{Cat}_1\in\mathbb{R}^{p_{OH}^{Cat}\times1}$ and GELU activation function following
\begin{align*}
\bm{E}^{Cat}_{n,1}(\bm{x}^{Cat}_{n})&=GELU(ELinear^{Cat}(\bm{v}_n))=GELU(\bm{W}^{Cat}_1\odot\bm{v}_n+\bm{b}^{Cat}_1)\\
&=(\bm{W}^{Cat}_1\odot\bm{v}_n+\bm{b}^{Cat}_1)\Phi(\bm{W}^{Cat}_1\odot\bm{v}_n+\bm{b}^{Cat}_1)
\end{align*}
where $\bm{E}^{Cat}_{n,1}$ is the first-layer categorical embeddings, $\Phi$ is the cumulative distribution function for the standard Gaussian distribution, and $\odot$ denotes the element-wise multiplication. 
The CatLinear layer $CLinear^{Cat}$ consists of weights $\bm{W}^{Cat}_2\in\mathbb{R}^{p_{OH}^{Cat}\times1}$ and biases $\bm{b}^{Cat}_2\in\mathbb{R}^{p^{Cat}\times1}$. For $i$-th ($i=1,2,...,p^{Cat}$) categorical feature, the CatLinear layer can be expressed as
$$
CLinear^{Cat}_i[\bm{E}^{Cat}_{n,1}(\bm{x}^{Cat}_{n})R(i)]=[\bm{W}^{Cat}_{2}R(i)]^T[\bm{E}^{Cat}_{n,1}(\bm{x}^{Cat}_{n})R(i)]+\bm{b}^{Cat}_{2,i}
$$
and 
\begin{equation*}
\bm{E}^{Cat}_{n,2}(\bm{x}^{Cat}_{n})=[CLinear^{Cat}_1[\bm{E}^{Cat}_{n,1}(\bm{x}^{Cat}_{n})R(1)],..., CLinear^{Cat}_{p^{Cat}}[\bm{E}^{Cat}_{n,1}(\bm{x}^{Cat}_{n})R(p^{Cat})]]
\end{equation*}
where $\bm{E}^{Cat}_{n,2}$ is the second-layer categorical embeddings, $R(i)$ is a reduction matrix that selects $i$-th categorical feature embedding from the entire one-hot encoded embedding from the first-layer, and $\bm{b}^{Cat}_{2,i}$ denotes the bias parameter for $i$-th feature for second-layer. Thus, the categorical embedding $CatEm$ of $n$-th observation's categorical features $\bm{x}^{Cat}_n$ can be expressed as
$$
\bm{E}^{Cat}_{n,2}(\bm{x}^{Cat}_{n})=CatEm(\bm{x}^{Cat}_n)=CLinear^{Cat}\circ GELU\circ ELinear^{Cat}\circ OH(\bm{x}^{Cat}_n)\in\mathbb{R}^{p^{Cat}\times1}
$$
This design ensures that the dimensionality of the categorical embeddings matches that of the original categorical features, which facilitates direct control over the attention matrix and improves interpretability. For numerical features, two Element-Wise Linear layers, $ELinear^{Con, 1}$ ($\bm{W}^{Con}_1\in\mathbb{R}^{p^{Con}\times1}$ and $\bm{b}^{Con}_1\in\mathbb{R}^{p^{Con}\times1}$) and $ELinear^{Con, 2}$ ($\bm{W}^{Con}_2\in\mathbb{R}^{p^{Con}\times1}$ and $\bm{b}^{Con}_2\in\mathbb{R}^{p^{Con}\times1}$), with a GELU activation in between are used to project each continuous feature independently. Thus, the continuous embedding $ConEm$ of $n$-th observation's continuous features $\bm{x}^{Con}_n$ can be expressed as
$$
\bm{E}^{Con}_{n,2}(\bm{x}^{Con}_{n})=ConEm(\bm{x}^{Con}_n)=ELinear^{Con, 2}\circ GELU \circ ELinear^{Con, 1}(\bm{x}^{Con}_n)\in\mathbb{R}^{p^{Con}\times1}
$$
Consequently, the resulting feature embeddings for input features $\bm{x}_n$, which writes
$$
\bm{E}({\bm{x}_n})=[\bm{E}^{Cat}_{n,2}(\bm{x}^{Cat}_{n}),\bm{E}^{Con}_{n,2}(\bm{x}^{Con}_{n})]\in\mathbb{R}^{p\times1}
$$ 
preserve the original input dimensionality, and no feature dependencies are introduced at the embedding stage.

\textbf{Attention Architecture:} We modify the standard scaled dot-product attention in FCorrTransformer, as illustrated in Figure \ref{fig:FCorr_attention}. Unlike the standard attention mechanism, which applies independent linear projections to obtain the query, key, and value matrices, FCorrTransformer removes these projections entirely. Hence, query, key, and value matrices are the same, i.e., normalized feature embedding $\bm{E}^{LN}(\bm{x}_n)=LN\circ\bm{E}(\bm{x}_n)$ where $LN$ denotes the layer normalization, in FCorrTransformer. This design choice is motivated by the desire to prevent biased information from bypassing the attention mechanism and by the assumption that the embedding layers already provide sufficient representations for attention. The attention mechanism in the first encoder layer is therefore defined as
\begin{equation}
\text{Attention}(\bm{x}_n)=\bm{A}(\bm{x}_n)\bm{E}^{LN}(\bm{x}_n)=\text{SoftMax}(\dfrac{\bm{E}^{LN}(\bm{x}_n)\bm{E}^{LN}(\bm{x}_n)^T}{\sqrt{p}})\bm{E}^{LN}(\bm{x}_n)
\label{eq:attn}
\end{equation}
where $\bm{A}(\bm{x}_n)=\text{SoftMax}(\dfrac{\bm{E}^{LN}(\bm{x}_n)\bm{E}^{LN}(\bm{x}_n)^T}{\sqrt{p}})\in\mathbbm{R}^{p\times p}$ is the attention matrix. The output of the previous attention mechanism is then forwarded as the input of the subsequent encoder layer. Since each feature embedding is one-dimensional, the dot-product in the first attention layer effectively reduces to computing pairwise scalar similarities between features, which are then normalized by the SoftMax operator. Furthermore, since multi-head attention can be difficult to interpret and we believe that a single attention head is sufficient to capture the essential dependencies among features in tabular settings, we adopt a single-head attention design throughout this work.

\subsection{\texttt{CAR}: Counterfactual Attention Regularization}\label{subsec:CAR}

The architecture of the FCorrTransformer prevents any biased feature dependency from bypassing the attention matrix $\bm{A}$, thereby aligning fair predictions with fair attention matrices. 
We hypothesize the existence of a fair representation for sensitive features that remains informative enough for the model to preserve predictive performance while avoiding discrimination against any individual or group. 
To achieve such counterfactually fair attention, one of the intuitive solutions is counterfactual data augmentation (CDA), a technique established in prior literature \citep{kusnerCounterfactualFairness2017, mutluContrastiveCounterfactualFairness2022, maLearningCounterfactualFairness2023}. By augmenting the dataset with counterfactual pairs, identical in all features except for the sensitive features, we can isolate the specific influence of the sensitive feature within the resulting attention matrices. 
In the preliminary experiments, we have observed that directly regulating model predictions on counterfactual data can enforce counterfactual fairness. However, this approach substantially degrades predictive performance because the fairness constraint implicitly restricts not only the attention mechanism but also the downstream predictive network. As a result, the model's capacity to learn meaningful predictive patterns is significantly reduced. Alternatively, one may attempt to impose fairness constraints directly on the attention matrix by enforcing a ``fair'' representation. For example, in the case of a binary sensitive feature (e.g., male and female), the attention weights corresponding to the two groups can be constrained to their average. While this approach aims to eliminate biased dependencies at the attention level, it also leads to substantial performance deterioration. In particular, during the early stages of training, the imposed ``fair'' representation is subject to the artificial definition of fair attention (i.e., average), thereby disrupting the optimization process and hindering effective model training. To address these limitations, we adopt a softer regularization strategy that focuses specifically on the attention interactions between sensitive and non-sensitive features. 

For any given sample $\bm{x}_n$ comprising non-sensitive features $\bm{x}_r$ and a sensitive feature $x_s$, i.e., $\bm{x}_n=[\bm{x}_r, x_s]$, the CAR term is defined as
\begin{equation}
\mathcal{L}_{CAR}(\bm{x}_n)=\dfrac{1}{C^s}\sum_{i=1}^{C^s}\|\text{SoftMax}(\dfrac{\bm{E}^{LN}(\bm{x}_n)\bm{E}^{LN}(\bm{x}_n)^T}{\sqrt{p}})_{:,\sigma}-\text{SoftMax}(\dfrac{\bm{E}^{LN}(\bm{x}_r, s_i)\bm{E}^{LN}(\bm{x}_r, s_i)^T}{\sqrt{p}})_{:,\sigma}\|^2_2
\label{eq:car-cda}
\end{equation}
where $\sigma$ is the position of the sensitive feature. The CAR term is computed by permuting the sensitive feature across all unique sensitive categories $s_i$ for $i=1,2,...,C^s$. This $\sigma$-th column vector captures the dependencies between the sensitive feature and all non-sensitive features. In other words, $\mathcal{L}_{CAR}(\bm{x}_n)$ quantifies average discrepancy in attention weights between the original input and counterfactual permutations. It should be noted that only the elements in the first attention matrix admit the pairwise feature dependency. The input, i.e., attention scores, for subsequent layers are composition of all features, making the attention matrix unidentifiable from the feature dependency perspective. Thus full fairness intervention through CAR on deep attention layers becomes impossible.
The CAR objective minimizes the distributional variance of these dependency vectors across counterfactual sensitive categories and learns a fair embedding of the sensitive feature, thereby enforcing a group-invariant representation of the sensitive features in the attention space. 
However, this CDA-based approach introduces significant computational and memory overhead, particularly when dealing with multiple or high-dimensional sensitive features, as the framework computes the attention matrix $C^s+1$ times for each batch.

To mitigate this computational overhead, we adopt an optimization strategy inspired by AttenD \citep{gaciDebiasingPretrainedText2022}. During training, the input is augmented with all unique counterfactual sensitive categories, enabling the generation of all sensitive-related attention vectors within a single forward pass. This avoids the repetitive attention matrix computation in the CDA-based method. However, a significant challenge arises in the standard FCorrTransformer: if the model dimension is naively expanded according to the number of counterfactual sensitive categories (i.e., from $p$ features to $p + C^s$ features), the original sensitive feature and its augmented counterfactual categories are treated as distinct features with separate embedding parameters. Such a configuration is undesirable, as regularization applied to the naively expanded attention matrix no longer enforces fair feature dependencies across counterfactual realizations. Instead, the original sensitive feature and its augmented counterfactual categories may each learn distinct and biased embeddings while still minimizing the misspecified regularization term. Hence, the regularization operates on mismatched features, thereby losing its originally intended fairness implication. Our approach ensures that the sensitive feature's embeddings and linear projections are shared across all counterfactual categories. This design choice maintains the original parameter size while avoiding feature mismatch. Specifically, we introduce three specialized modules, \textit{SenLayerNorm}, \textit{SenElementWiseLinear}, and \textit{SenCatLinear}. These three modules extract corresponding parameters of the sensitive feature from their non-counterfactual counterparts to construct a unified representation for the augmented counterfactual categories.

\begin{figure}[!ht]
    \centering
    \begin{subfigure}[t]{0.49\textwidth}
        \centering
        \includegraphics[scale=0.29]{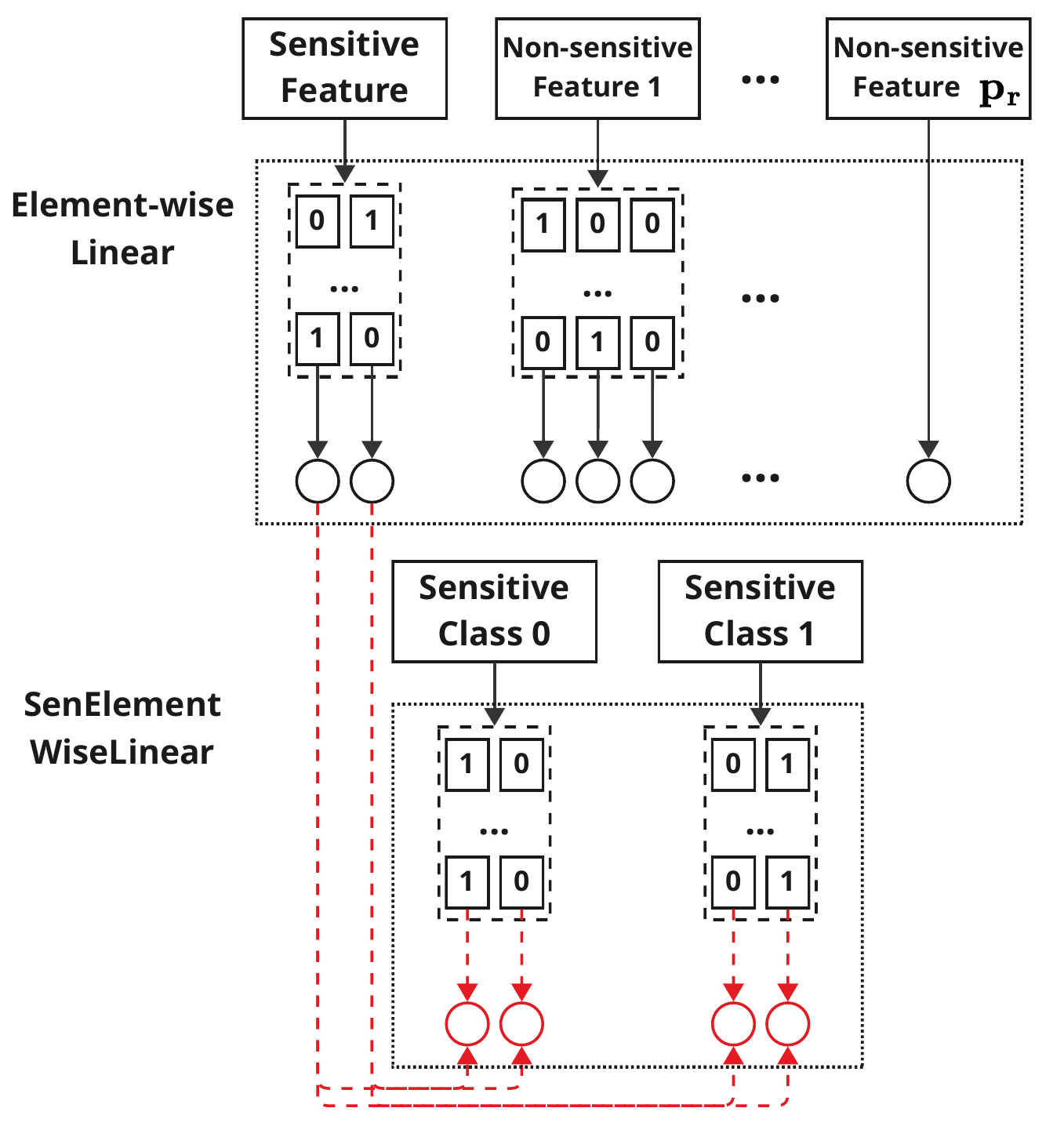}
        \caption{SenElementWiseLinear}
        \label{fig:SenElementWiseLinear}
    \end{subfigure}
    \begin{subfigure}[t]{0.49\textwidth}
        \centering
        \includegraphics[scale=0.3]{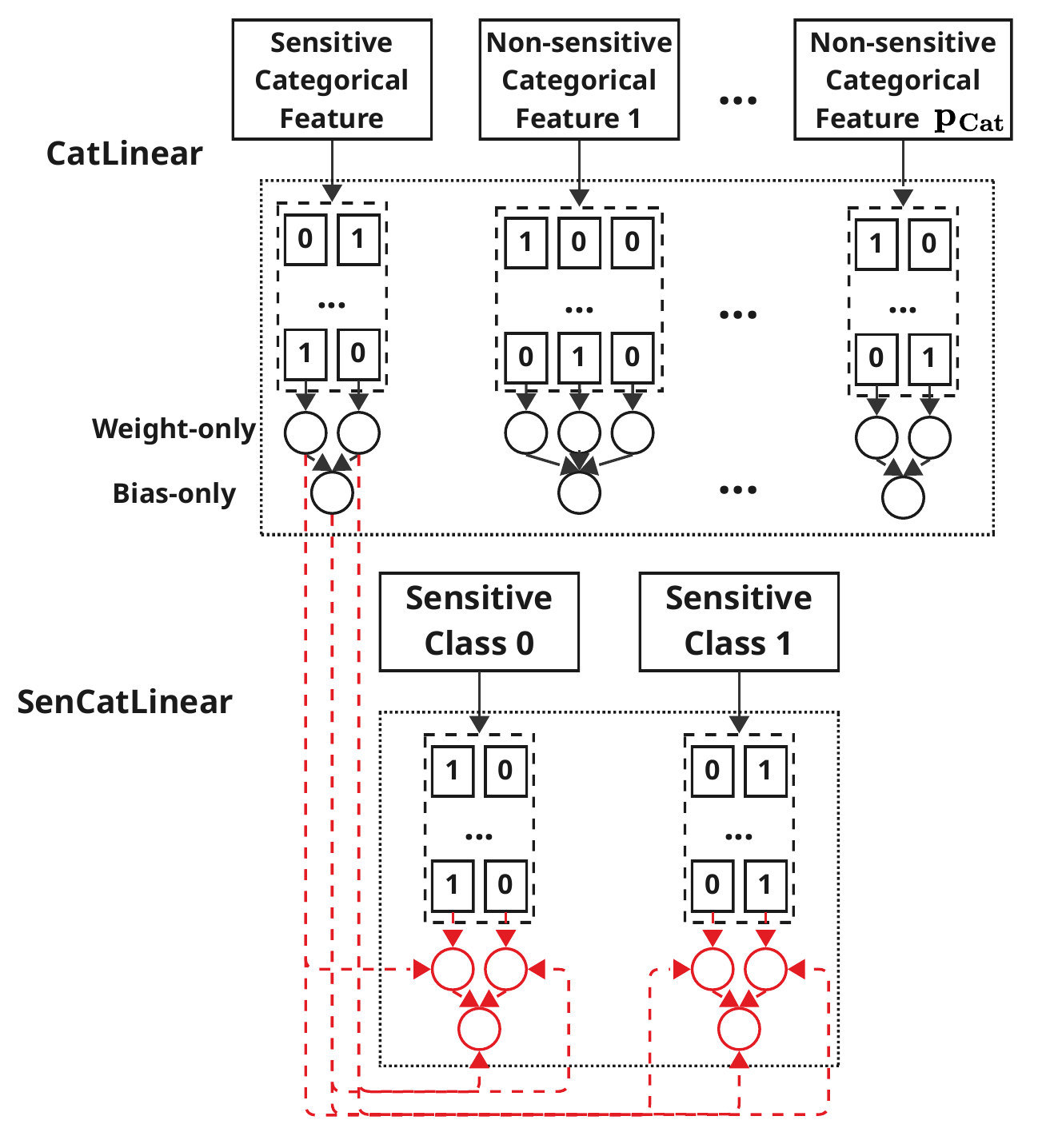}
        \caption{SenCatLinear}
        \label{fig:SenCatLinear}
    \end{subfigure}\\
    \caption{Illustration of SenElementWiseLinear and SenCatLinear on Binary Sensitive Feature}
    \label{fig:SenLayers}
\end{figure}

As illustrated in Figure \ref{fig:SenLayers}, during training, SenElementWiseLinear and SenCatLinear identify the indices of sensitive features and extract the corresponding weight and bias parameters from the standard Element-Wise Linear and CatLinear layers. These parameters and gradients are shared between the standard layers and the specialized modules. Thus, for input $\bm{x}_n\in\mathbb{R}^{p\times1}$, the augmented counterfactual sensitive categories $\bm{x}^{counter}_n=[s_1, s_2, ..., s_{C^s}]$. The corresponding SenElementWiseLinear layer for $i$-th senstive category $s_i$
\begin{align*}
SenELinear_i(OH(\bm{x}^{counter}_n))&=ELinear^{Cat}_{\sigma}[OH(\bm{x}^{counter}_n)R(i)]\\
&=[\bm{W}^{Cat}_1R(\sigma)]\odot [OH(\bm{x}^{counter}_n)R(i)]+\bm{b}^{Cat}_{1,\sigma}
\end{align*}
where $ELinear^{Cat}_{\sigma}$ is the standard Element-Wise Linear layer for the sensitive feature at index of $\sigma$ , $R$ denotes the reduction matrix and across all sensitive categories.
$$
SenELinear(OH(\bm{x}^{counter}_n))=[SenELinear_1(OH(\bm{x}^{counter}_n)), ..., SenELinear_{C^s}(OH(\bm{x}^{counter}_n))]
$$
We can then compute the first-layer sensitive categorical embeddings $\bm{E}^{Cat,Sen}_{n,1}(\bm{x}^{counter}_n)$ by applying the GELU activation function. Similarly, for SenCatLinear layers, the $i$-th sensitive category $s_i$
$$
SenCLinear_i[\bm{E}^{Cat,Sen}_{n,1}(\bm{x}^{counter}_n)]=CLiner^{Cat}_{\sigma}[\bm{E}^{Cat,Sen}_{n,1}(\bm{x}^{counter}_n)R(i)]
$$
and across all sensitive categories
$$
CatEm^{Sen}(\bm{x}^{counter}_n)=[SenCLinear_1[\bm{E}^{Cat,Sen}_{n,1}(\bm{x}^{counter}_n)], ..., SenCLinear_{C^s}[\bm{E}^{Cat,Sen}_{n,1}(\bm{x}^{counter}_n)]]
$$
Thus, the corresponding categorical embedding can be expressed as 
$$
\bm{E}^{Cat, Sen}_{n,2}(\bm{x}^{Cat}_{n})=[CatEm(\bm{x}^{Cat}_n), CatEm^{Sen}(\bm{x}^{counter}_n)]\in\mathbb{R}^{(p+C^S)\times1}
$$
where $CatEm(\bm{x}^{Cat}_n)$ follows standard categorical embedding. The overall feature embedding thus can be expressed as
$$
\bm{E}^{Sen}({\bm{x}_n})=[CatEm(\bm{x}^{Cat}_n), ConEm(\bm{x}^{Con}_n), CatEm^{Sen}(\bm{x}^{counter}_n)]\in\mathbb{R}^{(p+C^S)\times1}
$$ 

For SenLayerNorm, instead of normalizing over the augmented input space with the counterfactual categories ($p+C^s$), we compute the mean and standard deviation only over the original input ($p$). The corresponding weight and bias parameters of the sensitive features are extracted and applied to the augmented sensitive categories, ensuring consistency between input augmentation and CDA. For embedding $\bm{E}^{Sen}({\bm{x}_n})$ and a LayerNorm layer with weights $\bm{W}^{Norm}\in\mathbb{R}^{p\times1}$ and biases $\bm{b}^{Norm}\in\mathbb{R}^{p\times1}$, the SenLayerNorm can be expressed as 
\begin{equation*}
\medmath{
\bm{E}^{Sen, SLN}({\bm{x}_n})=[\dfrac{\bm{E}^{Sen}({\bm{x}_n})_{p+1}-\mathbb{E}[\bm{E}^{Sen}({\bm{x}_n})_{:p}]}{\sqrt{\text{Var}[\bm{E}^{Sen}({\bm{x}_n})_{:p}]}}W^{Norm}_{\sigma}+b^{Norm}_{\sigma}, ..., \dfrac{\bm{E}^{Sen}({\bm{x}_n})_{p+C^s}-\mathbb{E}[\bm{E}^{Sen}({\bm{x}_n})_{:p}]}{\sqrt{\text{Var}[\bm{E}^{Sen}({\bm{x}_n})_{:p}]}}W^{Norm}_{\sigma}+b^{Norm}_{\sigma}]
}
\end{equation*}
where $\mathbb{E}[\bm{E}^{Sen}({\bm{x}_n})_{:p}]=\dfrac{\sum_{j=1}^{p}\bm{E}^{Sen}({\bm{x}_n})_j}{p}$ and 
{\small
$\sqrt{\text{Var}[\bm{E}^{Sen}({\bm{x}_n})_{:p}]}=\sqrt{\dfrac{\sum_{j=1}^{p}(\bm{E}^{Sen}({\bm{x}_n})_j-\mathbb{E}[\bm{E}^{Sen}({\bm{x}_n})_{:p}])^2}{p}}$}
are the mean and standard deviation of the first $p$ features, and $W^{Norm}_{\sigma}$ and $b^{Norm}_{\sigma}$ are the $\sigma$-th (sensitive) feature weight and bias in the LayerNorm. It should be noted that the three sensitive layers are attached to their corresponding standard layers, and their outputs are concatenated before being passed to subsequent layers. Importantly, these sensitive layers reuse existing parameters and therefore introduce no additional learnable parameters.

For the attention mechanism and SoftMax with CAR, the dot-product can naturally handle expanded inputs since no linear projection is introduced. We retain the scale of $\sqrt{p}$ in dot-product operation, instead of the augmented space $\sqrt{p+C^s}$, to replicate the attention matrix in CDA-based method. The SoftMax is applied over the augmented space rather than the original feature space. In our prior experiments and analysis, SoftMax operation over the original feature space (i.e., first $p$ features) leads to attention weight explosion, especially in regression settings. As a result, attention weights under input augmentation are not strictly identical to those with CDA, but they converge to similar distributions after a few training iterations and at inference. After the FCorr encoder blocks, we discard the activations of the augmented sensitive categories and retain only the original feature activations as input to the prediction head.

Intuitively, this design reduces training time, as most of the computational cost in CDA arises from repeated attention computations over non-sensitive features. In our experiments introduced in Section \ref{sec:exp}, input augmentation reduces the average training time per epoch from 21.46 seconds to 19.50 seconds for classification and from 52.98 seconds to 14.40 seconds for regression under identical settings. The magnitude of this reduction is closely related to the dimensionality of the input and the cardinality of sensitive features. The classification task has low feature dimensionality, while the regression task has much higher dimensionality, leading to a more significant speedup. Overall, input augmentation provides a more efficient way to estimate variations in dependencies between sensitive and non-sensitive features. During inference, since input augmentation is disabled, this mechanism introduces no additional computational overhead. While our formulation and experiments focus on the single sensitive feature setting, the framework naturally extends to multiple sensitive features by augmenting all combinations of sensitive categories and applying the same parameter-sharing strategy. Our implementation fully supports the multi-sensitive-feature case. By expanding the feature space, the attention matrix is transformed from $\mathbb{R}^{p\times p}$ to $\mathbb{R}^{(p+C^s)\times(p+C^s)}$. The corresponding CAR term can be modified from Equation \ref{eq:car-cda} to
\begin{equation}
\medmath{
\mathcal{L}_{CAR}(\bm{x}_n)=\dfrac{1}{C^s}\sum_{i=1}^{C^s}\|\text{SoftMax}(\dfrac{\bm{E}^{Sen, SLN}(\bm{x}_n)\bm{E}^{Sen, SLN}(\bm{x}_n)^T}{\sqrt{p}})_{:p, \sigma}-\text{SoftMax}(\dfrac{\bm{E}^{Sen, SLN}(\bm{x}_n)\bm{E}^{Sen, SLN}(\bm{x}_n)^T}{\sqrt{p}})_{:p, p+i}\|_2^2 }
\label{eq:car-cia}
\end{equation}
where the subscript $:p,p+i$ denotes the first $p$ rows of the $(p+i)$-the column of the attention matrix with the augmented inputs. During inference, this augmentation process is suppressed to preserve computational efficiency, reverting the attention matrix to its original $p \times p$ dimensions.

During training, the CAR-incorporated loss function yields the following objective:
$$
\mathcal{L}=\mathcal{L}_{perf}+\lambda\mathcal{L}_{CAR}
$$
where $\mathcal{L}_{perf}$ denotes the model performance loss, $\lambda$ is regularization coefficient. In practice, we set the regularization coefficient as $\lambda=10^{\lfloor\mathcal{L}_{perf}^1/\mathcal{L}_{CAR}^1\rfloor}$, where $\mathcal{L}_{perf}^1$ and $\mathcal{L}_{CAR}^1$ denotes the corresponding losses on the first batch sample and $\lfloor\cdot\rfloor$ is the floor function. The coefficient setting generally guarantees counterfactual fairness in our experiments. 
Given the regularization structure of \texttt{CAR}, the level of counterfactual fairness can be naturally controlled via the regularization coefficient. Detailed analyses on the controllability of counterfactual fairness are presented in Appendix \ref{appendix_sec:controllability}.
To maintain the structural integrity of the baseline architectures, we employ the standard CDA-based approach for all comparative models, while utilizing the optimized input augmentation method for the FCorrTransformer.

\section{Experiments}\label{sec:exp}

To evaluate the proposed framework, we first conduct experiments on a synthetic toy dataset in Subsection \ref{subsec:sim} to analyze the attention mechanism in FCorrTransformer and to investigate how CAR influences the learning process. We then evaluate the framework on two imbalanced datasets in Subsection \ref{subsec:bak} and \ref{subsec:insurtech}, motivated by applications in domains such as insurance and finance, where strong fairness constraints are mandated by regulators and data distributions are extremely skewed. Since the proposed framework is applicable to both regression and classification settings, we evaluate its performance under both tasks.

\subsection{Synthetic Data}\label{subsec:sim}

To understand the feature dependency in the FCorrTransformer and the mechanism of CAR, we construct synthetic data following 
$$
X_1\sim
\begin{cases}
    Ber(0.7),\quad\text{if}X_2=1\\
    Ber(0.3),\quad\text{if}X_2=0
\end{cases}
X_2\sim Ber(0.5)\quad
X_3\sim Ber(0.5)\quad
y\sim
\begin{cases}
    Ber(0.8),\quad\text{if}X_2+X_3=2\\
    Ber(0.2),\quad\text{o.w}\\
\end{cases}
$$
where $X_2$ and $X_3$ are the informative features, $X_1$ is a sensitive feature correlated with $X_2$, and $Ber$ is Bernoulli distribution. Consequently, $X_1$ affects $y$ indirectly through $X_2$, which corresponds to indirect discrimination in the algorithmic fairness literature. A similar indirect discrimination data structure has been studied in \citet{mehrabiAttributingFairDecisions2022}. We train both FCorrTransformer and FCorrTransformer with CAR on this dataset. For counterfactual fairness evaluation, we adopt the Pairwise Comparison Metric (PCM) framework summarized in \citet{czarnowskaQuantifyingSocialBiases2021}, which can be formulated as
$$
PCM_{counterfactual}(\mathcal{D})=\dfrac{1}{\zeta}\sum_{s_{i}\in\mathcal{S}}\sum_{\substack{s_{j},s_{k}\in\mathcal{S}\\ s_{j}\neq s_{k}}}\mu(\phi(\mathcal{D}'_{s_{i}\rightarrow s_{j}}), \phi(\mathcal{D}'_{s_{i}\rightarrow s_{k}}))
$$
where $\mu$ and $\phi$ denotes certain distance and score function to evaluate the fairness, $\zeta$ is the normalization factor, $\mathcal{D}'$ is the perturbed data given the input $\mathcal{D}$ by perturbing all the sensitive features, and the data partition $\mathcal{D}'_{s_{i}\rightarrow s_{j}}$ replaces its sensitive feature from $s_{i}$ to perturbed category $s_{j}$. 
By iteratively perturb every sensitive features, $\forall i, j$, the original dataset $\mathcal{D}$ changed to $\mathcal{D}'=\cup_{s_{i},s_{j}\in\mathcal{S}}\mathcal{D}'_{s_{i}\rightarrow s_{j}}$.\footnote{To help understand the data perturbation process, let us consider a 100-sample dataset $\mathcal{D}$ with an univariate binary sensitive feature of gender, with 40 males (category $s_1$) and 60 females (category $s_2$). The perturbed data partition $\mathcal{D}'_{s_1\rightarrow s_1}$ is identical to the 40 males in $\mathcal{D}$, and the partition  $\mathcal{D}'_{s_1\rightarrow s_2}$ refers to 40 counterfactual ``females" that are otherwise identical to the 40 males (except for the sensitive feature, gender). We can also apply similar perturbation on the females to get $\mathcal{D}'_{s_2\rightarrow s_1}$ (60 counterfactual males) and $\mathcal{D}'_{s_2\rightarrow s_2}$ (60 original females), and the perturbed data $\mathcal{D}'$ is the combination of the four partitions with total sample size equals 200.}
For the classification tasks, we adopt the Wasserstein distance along with several standard performance metrics as score functions, formulating Average Individual Fairness (AvgIF) \citep {huangReducingSentimentBias2020} and F1/AUROC/AUPRC Gap. Please refer to Appendix \ref{appendix_sec:performance-fair} for the formulation of the fairness metrics. As shown in Table \ref{tab:fairness-sim}, evaluation on the synthetic dataset indicates that incorporating CAR enables the model to learn fair representations and effectively remove bias.

\begin{table}[!ht]
\centering
\caption{Fairness Metrics Comparison on Synthetic Data}
\label{tab:fairness-sim}
\begin{tabular}{ccccc}
\toprule
Model & AvgIF & F1 Gap & AUROC Gap & AUPRC Gap \\
\hline
FCorrTransformer & 0.0085 & \textbf{0.0000} & 0.0058 & 0.0021 \\
FCorrTransformer + CAR & \textbf{0.0000} & \textbf{0.0000} & \textbf{0.0000} & \textbf{0.0000} \\
\bottomrule
\end{tabular}
\end{table}

\begin{figure}[!ht]
    \centering
    \begin{subfigure}[t]{0.45\textwidth}
        \centering
        \includegraphics[scale=0.4]{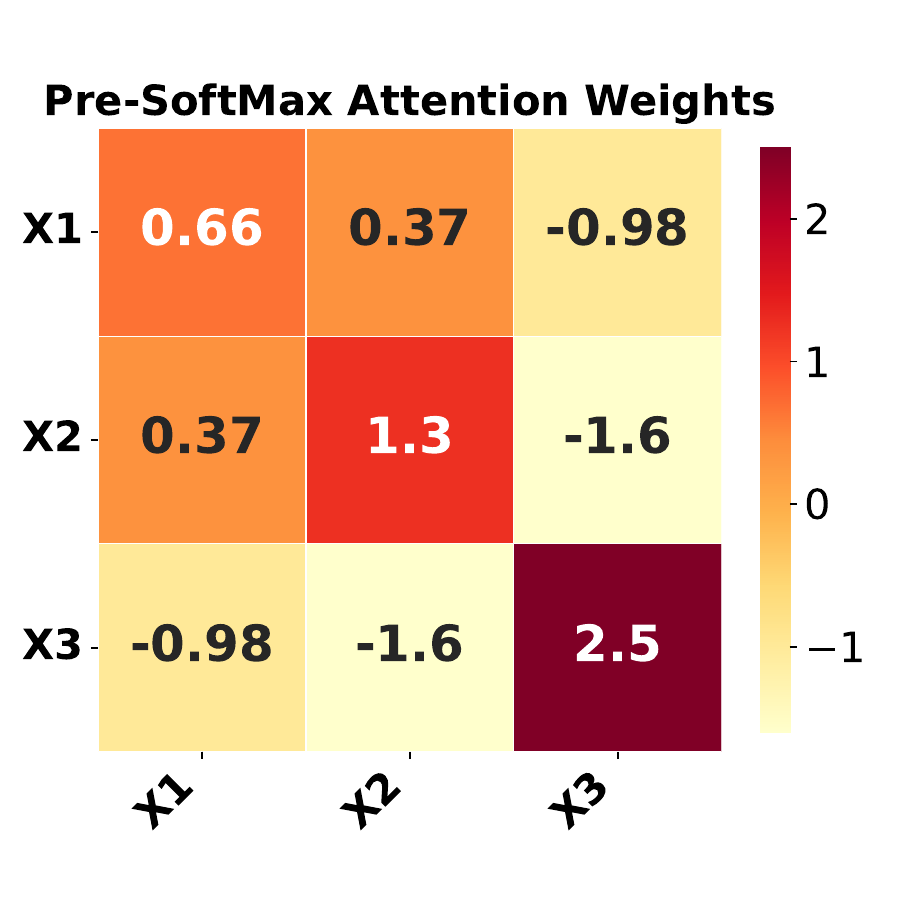}
        \caption{FCorrTransformer}
        \label{fig:attention_heatmap_fcorrtransformer_sim}
    \end{subfigure}
    \begin{subfigure}[t]{0.45\textwidth}
        \centering
        \includegraphics[scale=0.4]{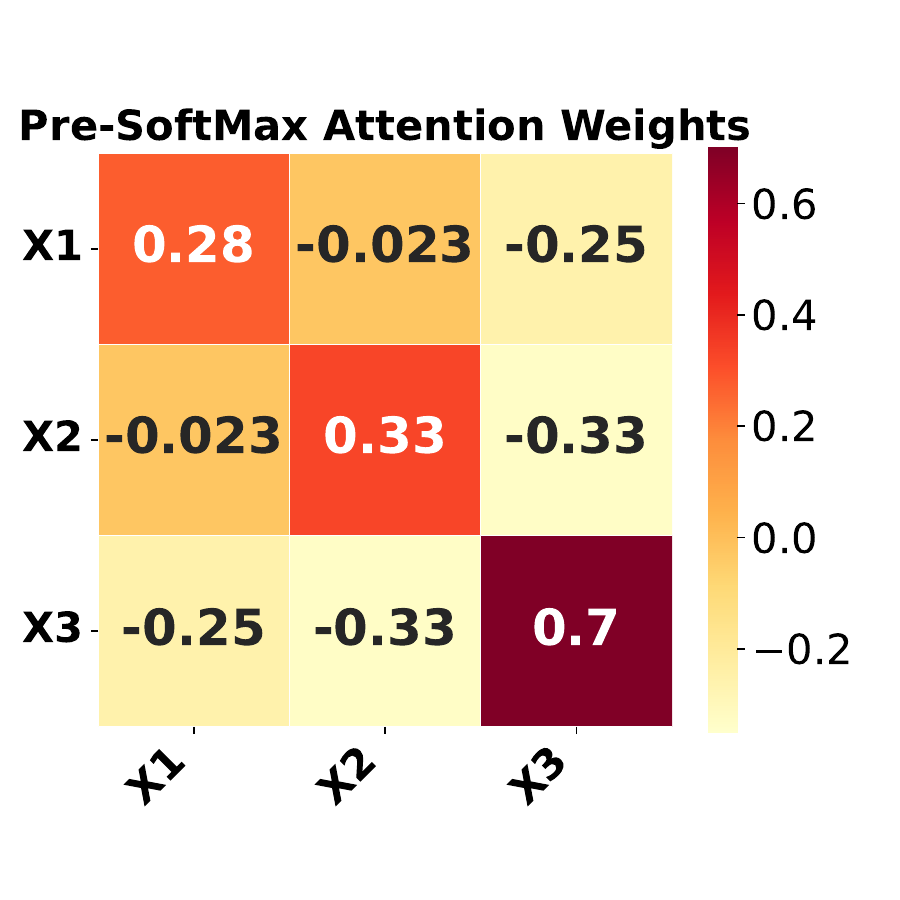}
        \caption{FCorrTransformer + CAR}
        \label{fig:attention_heatmap_fcorrtransformer_fair_sim}
    \end{subfigure}
    \caption{Heatmap of Pre-SoftMax Attention Weights on Synthetic Data}
    \label{fig:feature_attention_heatmap_sim}
\end{figure}

Figure \ref{fig:feature_attention_heatmap_sim} illustrates the pre-SoftMax attention matrix in FCorrTransformer and the effect of CAR. It is important to note that the attention matrix $\mathbf{A}(\mathbf{x}_n)$ in Equation~\ref{eq:attn} corresponds to the post-SoftMax matrix, which can be difficult to interpret due to the additional row-wise normalization. In contrast, the pre-SoftMax attention matrix reflects the learned pairwise feature dependencies, and takes values over the entire real line (as opposed to the $[0,1]$ range of the post-SoftMax matrix). Lower pre-SoftMax values correspond to lower post-SoftMax attention scores due to the exponential transformation in the SoftMax function. In the standard FCorrTransformer (Figure \ref{fig:attention_heatmap_fcorrtransformer_sim}), the diagonal elements, interpretable as feature significance, exhibit the largest values. In particular, the significance of $X_2$ and $X_3$ is 1.3 and 2.5, respectively, consistent with the data-generating process that these features are the informative features. The weak significance of $X_1$ (0.66) aligns with expectations, as $X_1$ is indirectly associated with $y$ through $X_2$. The off-diagonal elements capture pairwise feature dependencies. Notably, the interaction between $X_1$ and $X_2$ has a value of 0.37, the largest among the off-diagonal entries, highlighting their strong dependency. In contrast, the interactions between $X_1$ and $X_3$, and between $X_2$ and $X_3$, are negative, indicating weak or negligible dependency, which is consistent with the data-generating design. With FCorrTransformer + CAR (Figure \ref{fig:attention_heatmap_fcorrtransformer_fair_sim}), the attention matrix generally exhibits smaller pre-SoftMax values due to the effect of regularization. In particular, the unintended biased interaction between $X_1$ and $X_2$ is substantially suppressed: the ratio of the interaction term to the self-attention of $X_2$ decreases from 0.28 (0.37/1.3) to 0.007 (0.023/0.33). Furthermore, the interaction between $X_1$ and $X_3$ remains low, reflecting their independence. 
Overall, these results demonstrate that CAR effectively mitigates unintended biased dependencies in the learned feature relationships, which contributes to the improved counterfactual fairness metrics reported in Table \ref{tab:fairness-sim}.

\subsection{Imbalanced Classification: BAF}\label{subsec:bak}

For the imbalanced classification task, we use the Bank Account Fraud (BAF) dataset, a tabular dataset generated via a Generative Adversarial Network (GAN) based on real-world fraud detection data \citep{jesusTurningTablesBiased2022}. We adopt the base variant of BAF\footnote{\url{https://www.kaggle.com/datasets/sgpjesus/bank-account-fraud-dataset-neurips-2022}}, which contains 1,000,000 samples and 31 features, with approximately 1.1\% labeled as fraudulent, reflecting the natural rarity of financial fraud in real-world settings. Following \citet{jesusTurningTablesBiased2022}, we select \textit{customer age} as the sensitive feature. This is a multi-class categorical feature representing age groups (e.g., 10s to 90s), with nine unique categories. As baseline models, we consider a standard feed-forward neural network (FFN) composed of fully connected layers, as well as attention-heavy tabular transformers, namely TabTransformer and FT-Transformer. In contrast, our proposed FCorrTransformer is an attention-light (FFN-heavy) architecture. The architecture and learning hyperparameters are optimized to maximize training Area Under the Receiver Operating Curve (AUROC). We further apply CAR to both FT-Transformer and FCorrTransformer to evaluate their performance and counterfactual fairness. The definitions of performance and fairness metrics are provided in Appendix \ref{appendix_sec:performance-metric} and \ref{appendix_sec:performance-fair}. In addition to counterfactual fairness metrics, we report standard group fairness measures, including Demographic Parity Difference (DPD), Equalized Odds (EqOdd) \citep{agarwalReductionsApproachFair2018}, and Equal Opportunity (EqOpp) \citep{hardtEqualityOpportunitySupervised2016}. The performance and fairness results are summarized in Tables \ref{tab:performance-BAF} and \ref{tab:fairness-BAF}.

\begin{table}[!ht]
\centering
\caption{Model Performance Comparison on BAF dataset}
\label{tab:performance-BAF}
\footnotesize
\begin{tabular}{cccccccc}
\toprule
Subset & Model & Accuracy & F1 score & FPR & FNR & AUROC & AUPRC \\
\hline
\multirow{6}*{Train} & FFN & \textbf{0.9880} & 0.1286 & \textbf{0.0017} & 0.9209 & 0.8676 & 0.1342 \\
& TabTransformer & 0.9775 & 0.2096 & 0.0144 & 0.7330 & 0.8721 & 0.1370 \\
& FT-Transformer & 0.9842 & \textbf{0.2444} & 0.0073 & 0.7702 & \textbf{0.8988} & \textbf{0.1803} \\
& FT-Transformer + CAR & 0.9833 & 0.2392 & 0.0082 & 0.7650 & 0.8955 & 0.1714 \\
& FCorrTransformer & 0.9824 & 0.2343 & 0.0093 & 0.7578 & 0.8928 & 0.1648 \\
& FCorrTransformer + CAR &  0.9794 & 0.2324 & 0.0127 & \textbf{0.7202} & 0.8912 & 0.1576 \\
\hline
\multirow{6}*{Test} & FFN & \textbf{0.9883} & 0.1148 & \textbf{0.0020} & 0.9279 & 0.8683 & 0.1234 \\
& TabTransformer & 0.9773 & 0.1871 & 0.0149 & 0.7525 & 0.8635 & 0.1080 \\
& FT-Transformer & 0.9844 & \textbf{0.2342} & 0.0076 & 0.7734 & \textbf{0.8955} & \textbf{0.1607} \\
& FT-Transformer + CAR & 0.9834 & 0.2268 & 0.0085 & 0.7696 & 0.8945 & 0.1602 \\
& FCorrTransformer & 0.9826 & 0.2220 & 0.0094 & 0.7648 & 0.8925 & 0.1458 \\
& FCorrTransformer + CAR & 0.9796 & 0.2230 & 0.0130 & \textbf{0.7217} & 0.8910 & 0.1453 \\
\bottomrule
\end{tabular}
\end{table}

\begin{table}[!ht]
\centering
\caption{Fairness Metrics Comparison on BAF dataset}
\label{tab:fairness-BAF}
\scriptsize
\begin{tabular}{ccccccccc}
\toprule
Subset & Model & DPD & EqOdd & EqOpp & AvgIF & F1 Gap & AUROC Gap & AUPRC Gap \\
\hline
\multirow{6}*{Train} & FFN & 0.0462 & 0.2500 & 0.2500 & 0.0717 & 0.7539 & 0.0441 & 0.1152 \\
& TabTransformer & 0.1196 & 0.5145 & 0.5145 & 0.1685 & 0.5458 & 0.1049 & 0.2065 \\
& FT-Transformer & 0.0928 & 0.5286 &  0.5286 & 0.0835 & 0.4953 & 0.0244 & 0.0667 \\
& FT-Transformer + CAR & 0.0752 & 0.4432 & 0.4432 & 0.0629 & 0.4084 & 0.0080 & 0.0234 \\
& FCorrTransformer & 0.1227 & 0.5999 & 0.5999 & 0.0747 & 0.4153 & 0.0215 & 0.1920 \\
& FCorrTransformer + CAR & \textbf{0.0456} & \textbf{0.2325} & \textbf{0.2325} & \textbf{0.0000} & \textbf{0.0000} & \textbf{0.0000} & \textbf{0.0000} \\
\hline
\multirow{6}*{Test} & FFN & \textbf{0.0391} & \textbf{0.5714} & \textbf{0.5714} & 0.0675 & 0.6389 &   0.0922 & 0.0531 \\
& TabTransformer & 0.1200 & 0.8571 & 0.8571 & 0.1691 & 0.3366 & 0.0509 & 0.1108 \\
& FT-Transformer & 0.0969 & 0.8571 & 0.8571 & 0.0796 & 0.2347 & 0.0135 & 0.0545 \\
& FT-Transformer + CAR & 0.0904 & 0.8571 & 0.8571 & 0.0647 & 0.2816 & 0.0029 & 0.0151 \\
& FCorrTransformer & 0.1004 & 0.8571 &  0.8571 & 0.0734 & 0.2148 & 0.0160 & 0.0805 \\
& FCorrTransformer + CAR & 0.0859 & 0.7143 & 0.7143 & \textbf{0.0000} & \textbf{0.0000} & \textbf{0.0000} & \textbf{0.0000} \\
\bottomrule
\end{tabular}
\end{table}

Table \ref{tab:performance-BAF} shows that FT-Transformer and FCorrTransformer outperform FFN and TabTransformer in terms of F1 score, AUROC, and Area Under Precision Recall Curve (AUPRC). In contrast, metrics such as Accuracy, False Positive Rate (FPR), and False Negative Rate (FNR) are highly sensitive to the decision threshold, which is determined independently as discussed in Appendix \ref{appendix_subsec:bak}, and are therefore less informative in highly imbalanced classification settings. FCorrTransformer exhibits a noticeable performance gap compared to FT-Transformer, which is expected due to its simpler architectural design. As discussed in Appendix \ref{appendix_subsec:bak}, this gap can be partially mitigated by reintroducing residual connections; however, doing so weakens the model’s fairness guarantees. When applied, CAR introduces a noticeable performance drop for both FT-Transformer and FCorrTransformer, while some of the threshold-sensitive metrics do not suffer significant performance degradation. More importantly, the fairness results in Table \ref{tab:fairness-BAF} reveal substantial variation across models. While CAR generally reduces bias in FT-Transformer across all fairness metrics, significant residual bias remains. For FCorrTransformer, although complete group fairness is not guaranteed, since group fairness is inherently data-dependent and not explicitly enforced by CAR, the combination of FCorrTransformer + CAR achieves strong counterfactual fairness with only minimal performance loss. This property is particularly valuable in applications where strict fairness is required by regulatory standards. It should be noted that, coincidentally, the optimized FCorrTransformer architectures in both the BAF experiment and the subsequent imbalanced regression experiments are single encoder layer FCorrTransformer, which admits different interpretation for the feature attention matrices compared to multi-layer ones. Our discussion in Appendix \ref{appendix_sec:multi-layer} suggests that CAR can still achieve counterfactual fairness in the multi-layer architectures.

\begin{figure}[!ht]
    \centering
    \begin{subfigure}[t]{\textwidth}
        \centering
        \includegraphics[scale=0.35]{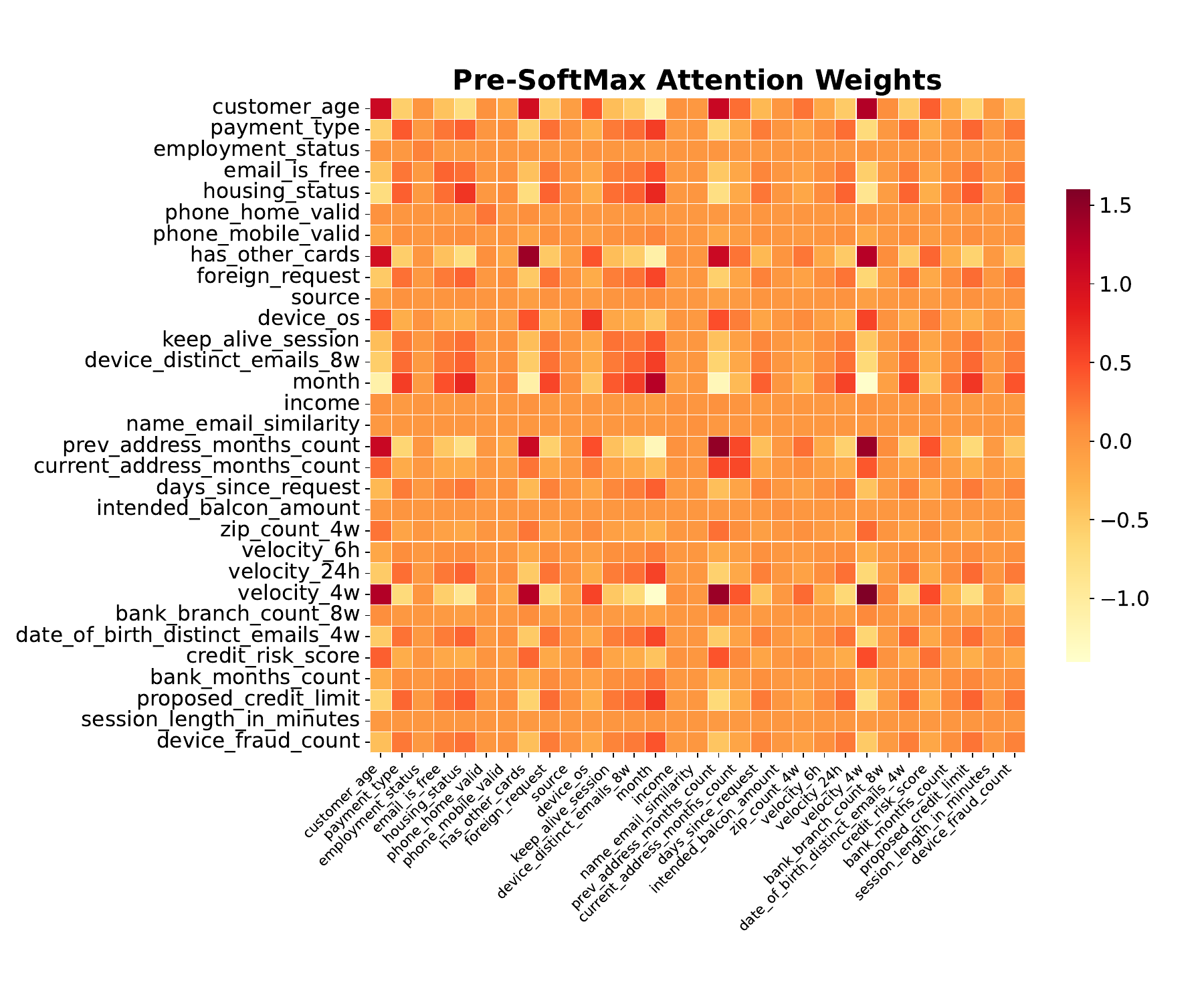}
        \caption{FCorrTransformer}
        \label{fig:attention_heatmap_fcorrtransformer}
    \end{subfigure}
    \begin{subfigure}[t]{\textwidth}
        \centering
        \includegraphics[scale=0.35]{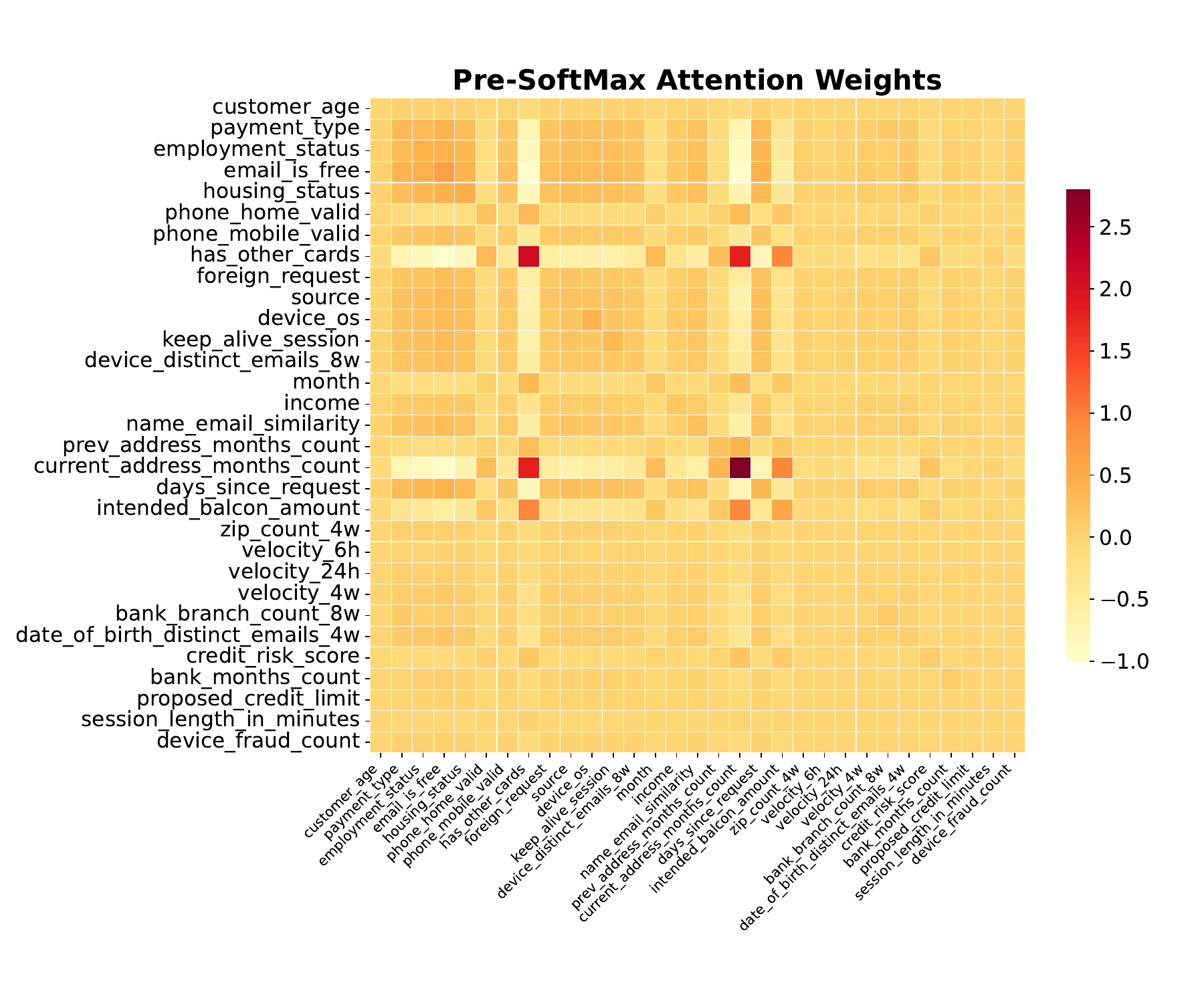}
        \caption{FCorrTransformer + CAR}
        \label{fig:attention_heatmap_fcorrtransformer_fair}
    \end{subfigure}\\
    \caption{Heatmap of Pre-SoftMax Attention Weights on BAF dataset}
    \label{fig:feature_attention_heatmap_BAF}
\end{figure}

\begin{figure}[!ht]
    \begin{subfigure}[t]{\textwidth}
    \centering
    \includegraphics[scale=0.4]{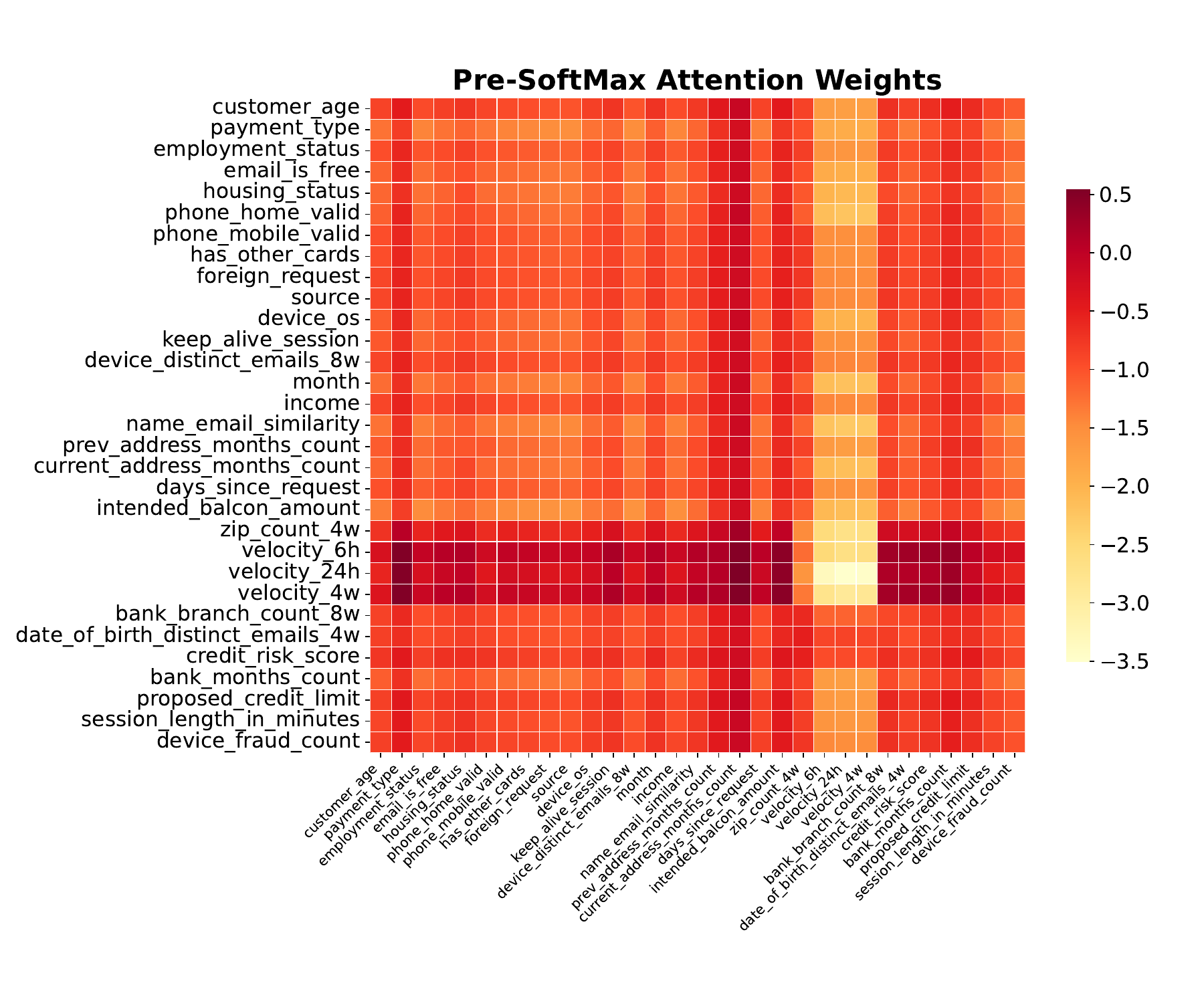}
    \caption{Feature attention weights}
    \label{fig:attention_heatmap_fttransformer}
    \end{subfigure}
    \begin{subfigure}[t]{\textwidth}
    \centering
    \includegraphics[scale=0.4]{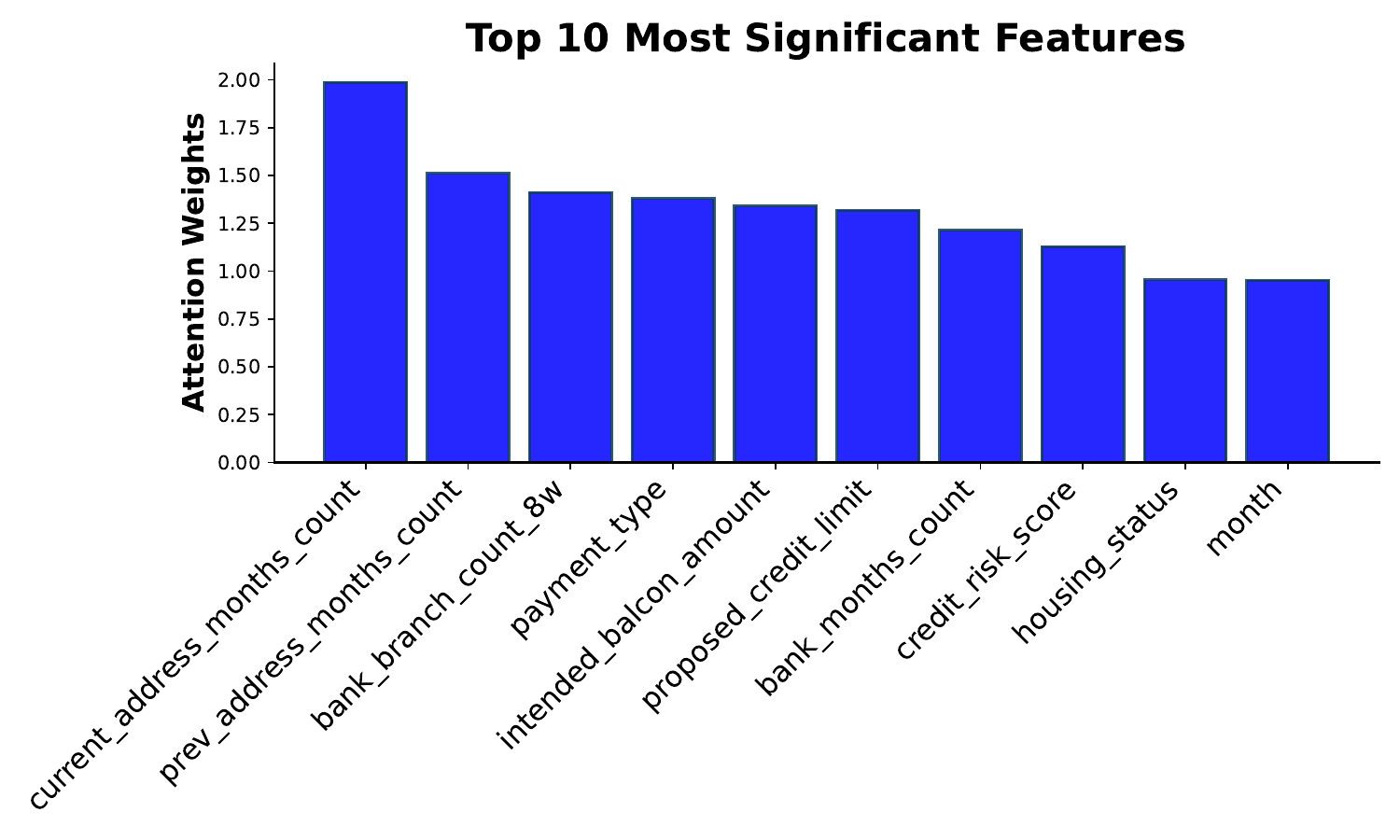}
    \caption{Top 10 feature significance}
    \label{fig:top_10_feature_importance_fttransformer}
    \end{subfigure}
    \caption{Attention Weights and Top 10 Features in FT-Transformer on BAF dataset}
    \label{fig:fttransformer_BAF}
\end{figure}

To better understand the mechanism of CAR, Figure \ref{fig:feature_attention_heatmap_BAF} visualizes the average pre-SoftMax attention matrices over the training set. In contrast to the classical attention mechanism used in FT-Transformer (Figure \ref{fig:attention_heatmap_fttransformer}), whose attention weights lack clear statistical interpretation, the attention patterns in FCorrTransformer are more structured and closely resemble feature dependencies. As shown in Figure \ref{fig:attention_heatmap_fcorrtransformer}, diagonal elements can be interpreted as the significance of individual features, while off-diagonal elements capture pairwise feature dependencies. For the purpose of visualization, the sensitive feature is placed in the first position in all experiments. The attention matrix indicates that the sensitive feature, customer age, receives high self-attention and exhibits strong dependencies with several informative features, enabling the model to exploit sensitive information. As discussed in Subsection \ref{subsec:sim}, CAR suppresses unintended biased feature dependencies and reallocates attention toward fair features. This effect is clearly illustrated in Figure \ref{fig:attention_heatmap_fcorrtransformer_fair}, where attention to the sensitive feature is substantially suppressed. 

\begin{figure}[!ht]
    \centering
    \begin{subfigure}[t]{\textwidth}
        \centering
        \includegraphics[scale=0.4]{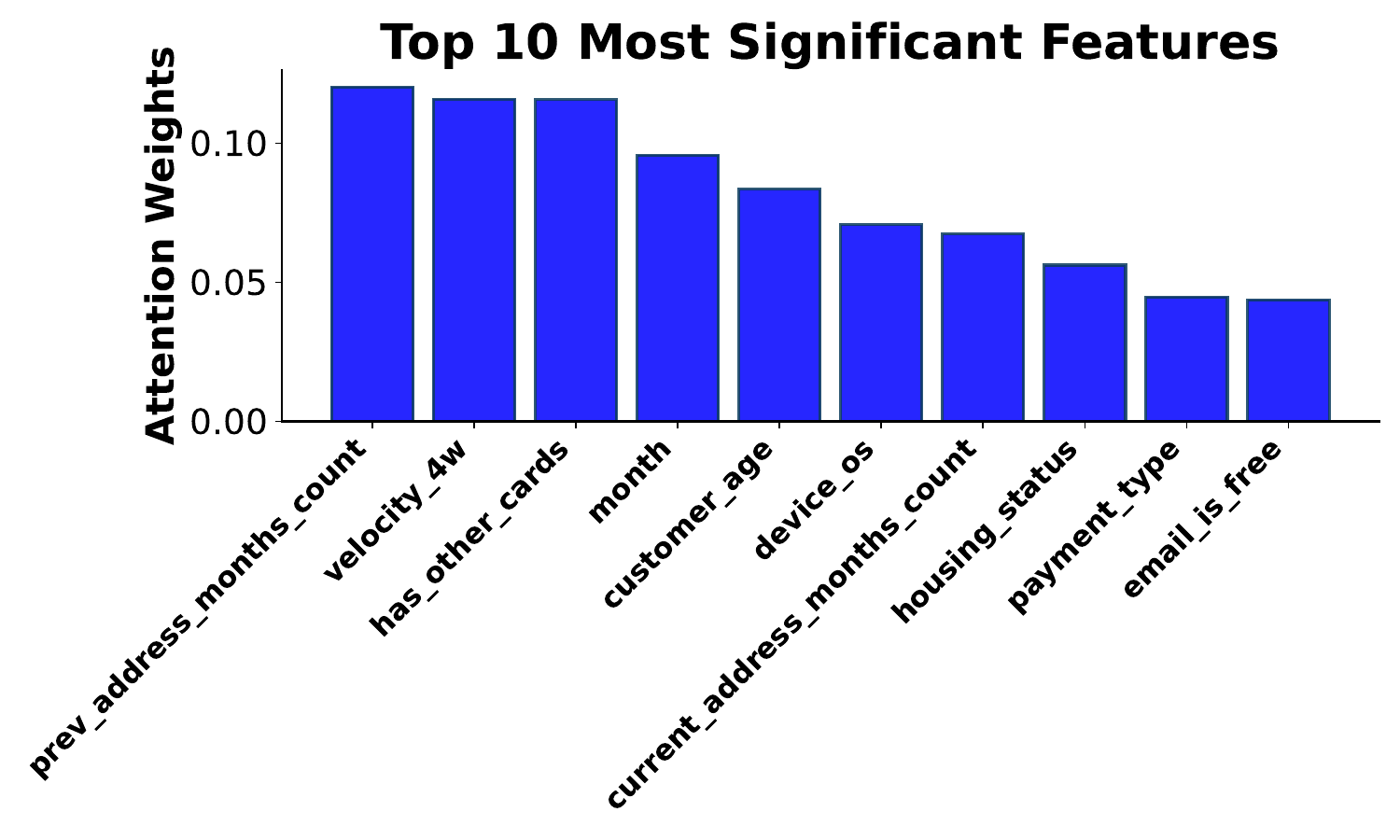}
        \caption{FCorrTransformer}
        \label{fig:top_10_feature_importance_fcorrtransformer}
    \end{subfigure}
    \begin{subfigure}[t]{\textwidth}
        \centering
        \includegraphics[scale=0.4]{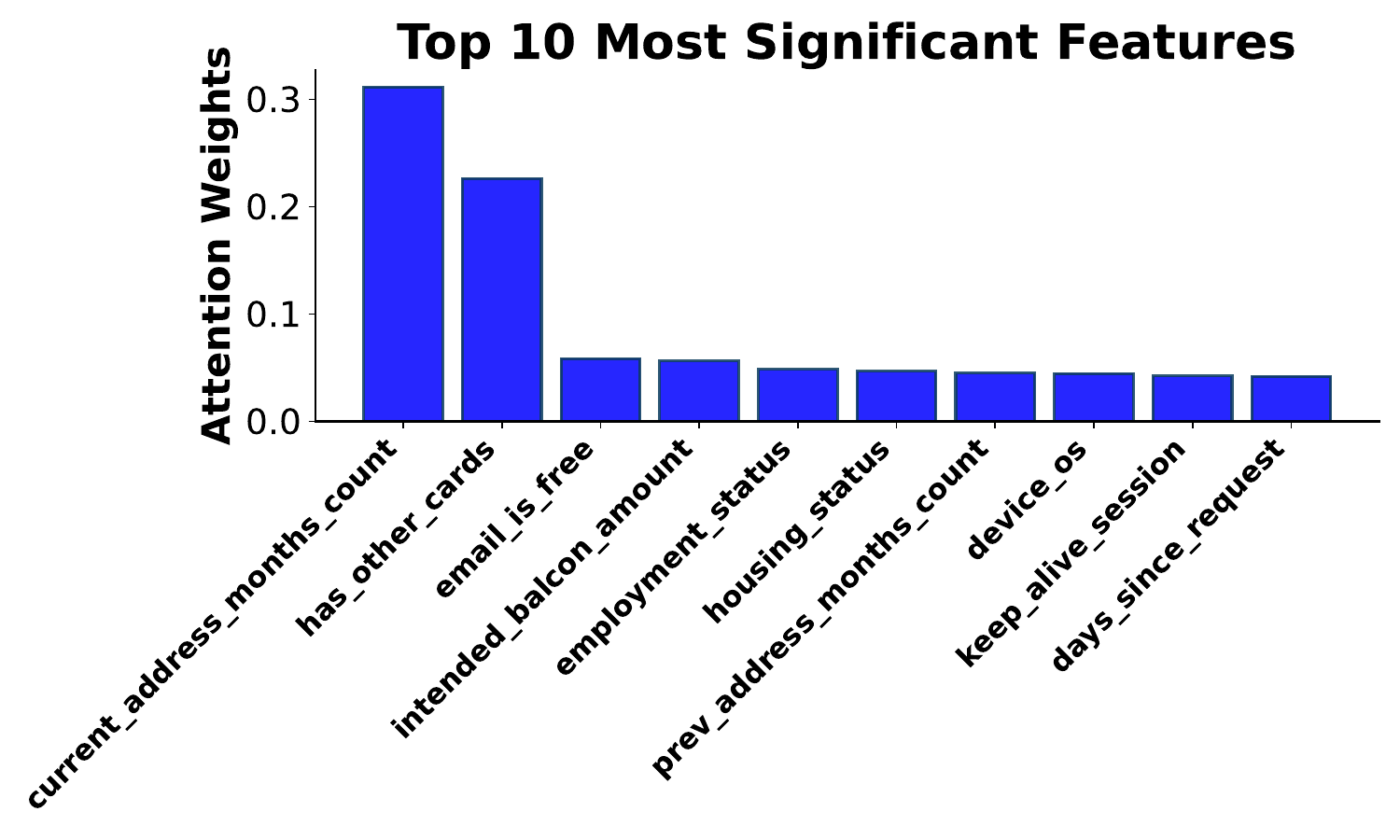}
        \caption{FCorrTransformer + CAR}
        \label{fig:top_10_feature_importance_fcorrtransformer_fair}
    \end{subfigure}
    \caption{Top 10 Feature Significance on BAF dataset}
    \label{fig:top_10_feature_importance_BAF}
\end{figure}

The CAR regularization centralizes attention on a few critical entries, such as the self-attention of \textit{current\_address\_months\_count} and \textit{has\_other\_cards}, as well as the dependencies between these two features, suggesting that they provide relatively fair predictive signals. In contrast, the attention on important features such as \textit{customer age}, \textit{velocity\_4w}, and \textit{prev\_address\_months\_count} is suppressed, as these features exhibit statistical dependencies with biased information. Overall, the attention mechanism in FCorrTransformer provides an interpretable perspective on how CAR reshapes feature dependencies during training and how such dependencies contribute to the model’s decision-making. Under this interpretation, diagonal attention weights naturally serve as a measure of feature significance. Figure \ref{fig:top_10_feature_importance_BAF} reports the top-10 most important features identified by FCorrTransformer, which largely agree with the feature importance extracted from LightGBM (Figure \ref{fig:top_10_feature_importance_lightgbm} in Appendix \ref{appendix_subsec:bak}), suggesting the feasibility of interpreting attention weights in FCorrTransformer as feature dependency.

For the attention matrix in FT-Transformer, as illustrated in Figure \ref{fig:attention_heatmap_fttransformer}, it exhibits a much denser attention distribution compared to FCorrTransformer. In contrast to FCorrTransformer, the statistical interpretation of the attention matrix in FT-Transformer is less clear. However, given the noticeable variation in attention weights across columns, we aggregate the attention weights by column (instead of using diagonal elements) and treat them as feature significance scores for FT-Transformer, as shown in Figure \ref{fig:top_10_feature_importance_fttransformer}. The resulting feature significance matches five of the features identified by LightGBM and shares most of the remaining features with FCorrTransformer. This suggests that FT-Transformer can still identify significant features, but its attention weights are less directly interpretable and provide limited insight into feature dependencies. In addition to the BAF dataset, we also evaluate the effectiveness of the FCorrTransformer and CAR on the well-studied Adult dataset in fairness literature. The performance and fairness metrics suggest similar findings observed in the BAF dataset. Please refer to Appendix \ref{appendix_subsec:adult} for detailed results on the Adult dataset.

\begin{table}[!ht]
\centering
\caption{Complexity of Each Model on BAF dataset}
\label{tab:mol-complexity-BAF}
\begin{tabular}{cccc}
\toprule
\multirow{2}*{Model} & \multirow{2}*{Number of parameters} & \multicolumn{2}{c}{GPU Memory Usage/MB (batch of 64)} \\
 &  & Parameter size & Forward/backward pass size \\
\hline
FFN & 74,897 & 0.30 & 0.40 \\
TabTransformer & 126,998 & 0.51 & 5.63 \\
FT-Transformer & 94,981 & 0.37 & 67.88 \\
FCorrTransformer & 11,064 & 0.04 & 0.25 \\
\bottomrule
\end{tabular}
\end{table}

Finally, due to its linear architecture and absence of high-dimensional embeddings, FCorrTransformer is computationally efficient. FCorrTransformer has significantly lower parameter count and memory usage compared to FT-Transformer as summarized in Table \ref{tab:mol-complexity-BAF}, enabling more efficient training and deployment. This efficiency stems from its attention-light design: the attention only consists 5.7\% of parameters in FCorrTransformer, while it accounts for 99.9\% of parameters in FT-Transformer.

\subsection{Imbalanced Regression: InsurTech}\label{subsec:insurtech}

We further evaluate the proposed framework on an imbalanced regression task using the InsurTech dataset introduced by \citet{quanImprovingBusinessInsurance2025}. This dataset consists of real-world Business Owner’s Policy (BOP) insurance data enriched with InsurTech features extracted from online sources. We focus on the liability (LIAB) coverage line, which includes 235,636 policies and 581 features in the train/test split. The response feature is the claim amount incurred during the policy period. Consistent with the characteristics of business liability insurance, the data exhibit extreme imbalance: only 1,870 policies (approximately 0.79\%) report non-zero claims, and the positive claim amounts are highly skewed, with a mean of 27,507.90 and a standard deviation of 33,244.72. Please refer \citet{quanImprovingBusinessInsurance2025} for more details on the dataset. We select business credit score as the sensitive feature, which is a multi-class categorical feature with seven categories (A+, A, B+, B, C+, C, and missing). The models are optimized to minimize Mean Squared Error (MSE). 

\begin{table}[!ht]
\centering
\caption{Model Performance Comparison on InsurTech dataset}
\label{tab:performance-InsurTech}
\begin{tabular}{cccccc}
\toprule
Subset & Model & Gini & PE & RMSE & MAE \\
\hline
\multirow{5}*{Train} & LightGBM & \textbf{0.78} & \textbf{0.00} & \textbf{3,853.67} & \textbf{435.65} \\
& FT-Transformer & 0.52 & -0.03 & 3,927.13 & 453.69 \\
& FT-Transformer + CAR & 0.50 & 0.01 & 3,928.28 & 444.24 \\
& FCorrTransformer & 0.57 & 0.01 & 3,915.64 & 441.94 \\
& FCorrTransformer + CAR & 0.54 & -0.01 & 3,918.32 & 445.86 \\
\hline
\multirow{5}*{Test} & LightGBM & \textbf{0.56} & \textbf{-0.26} & \textbf{3,305.85} & \textbf{394.56} \\
& FT-Transformer & 0.46 & -0.32 & 3,311.66 & 407.21 \\
& FT-Transformer + CAR & 0.46 & \textbf{-0.26} & 3,311.97 & 398.05 \\
& FCorrTransformer & 0.54 & \textbf{-0.26} & 3,306.11 & 396.06 \\
& FCorrTransformer + CAR & 0.50 & -0.28 & 3,310.45 & 398.99 \\
\bottomrule
\end{tabular}
\end{table}

For the InsurTech experiment, we report the same performance metrics used in \citet{quanImprovingBusinessInsurance2025}, which capture prediction ranking, portfolio-level performance, and individual prediction accuracy, key criteria for practical decision-making in insurance applications. Note that we make two minor modifications to the dataset compared to the original LightGBM setup. First, since neural networks cannot inherently handle missing values, while LightGBM can, we apply simple imputation using the mean for continuous features and the mode for categorical features. While this treatment is not optimal, it is a simplified approach strictly for comparative purposes for neural network-based approaches. Second, as all multi-class categorical features are binarized in the original dataset (e.g., separate indicators for credit scores A+, A, etc.), we regroup the credit score indicators into a single seven-class categorical feature. In our experiments, straightforward FFN collapses to a constant prediction (202.4047) despite extensive optimization. While this result achieves absolute fairness, it lacks practical utility for real-world insurance applications, indicating that simple linear architectures struggle to capture signal within complex, imbalanced insurance datasets. Table \ref{tab:performance-InsurTech} summarizes the performance of all evaluated models. Overall, LightGBM remains the most stable and best-performing model in our experiments. However, FCorrTransformer generally outperforms FT-Transformer despite its simpler architecture and smaller parameter size, suggesting that preserving raw feature values, rather than relying on high-dimensional embeddings, can be critical for modeling complex real-world tabular data.

\begin{table}[!ht]
\centering
\caption{Fairness Metrics Comparison on InsurTech dataset}
\label{tab:fairness-InsurTech}
\begin{tabular}{cccccc}
\toprule
Subset & Model & DPD & AvgIF & RMSE Gap & MAE Gap \\
\hline
\multirow{4}*{Train} & FT-Transformer & 135.98 & 187.15 & 7.22 & 182.27 \\
& FT-Transformer + CAR & 52.19 & 55.91 & 1.21 & 55.25 \\
& FCorrTransformer & 75.82 & 90.74 & 4.11 & 88.46 \\
& FCorrTransformer + CAR & \textbf{41.48} & \textbf{0.00} & \textbf{0.00} & \textbf{0.00} \\
\hline
\multirow{4}*{Test} & FT-Transformer & 140.32 & 187.49 & 32.63 & 181.33 \\
& FT-Transformer + CAR & \textbf{53.03} & 55.97 & 9.38 & 55.39 \\
& FCorrTransformer & 113.62 & 92.76 & 19.23 & 90.51 \\
& FCorrTransformer + CAR & 59.47 & \textbf{0.00} & \textbf{0.00} & \textbf{0.00} \\
\bottomrule
\end{tabular}
\end{table}

Furthermore, the fairness evaluations in Table \ref{tab:fairness-InsurTech} reveal that CAR significantly enhances both group-level and counterfactual fairness across FT-Transformer and FCorrTransformer. In particular, only the FCorrTransformer combined with CAR provides robust counterfactual fairness guarantees that are consistent with stringent regulatory expectations. This indicates that fairness cannot be achieved through regularization alone, but rather depends critically on the interaction between the model architecture and bias mitigation mechanisms. This suggests that carefully designed structures with complete feature dependency control is essential for enforcing strong fairness constraints in complex, high-dimensional tabular settings.

\begin{figure}[!ht]
    \centering
    \begin{subfigure}[t]{\textwidth}
        \centering
        \includegraphics[scale=0.4]{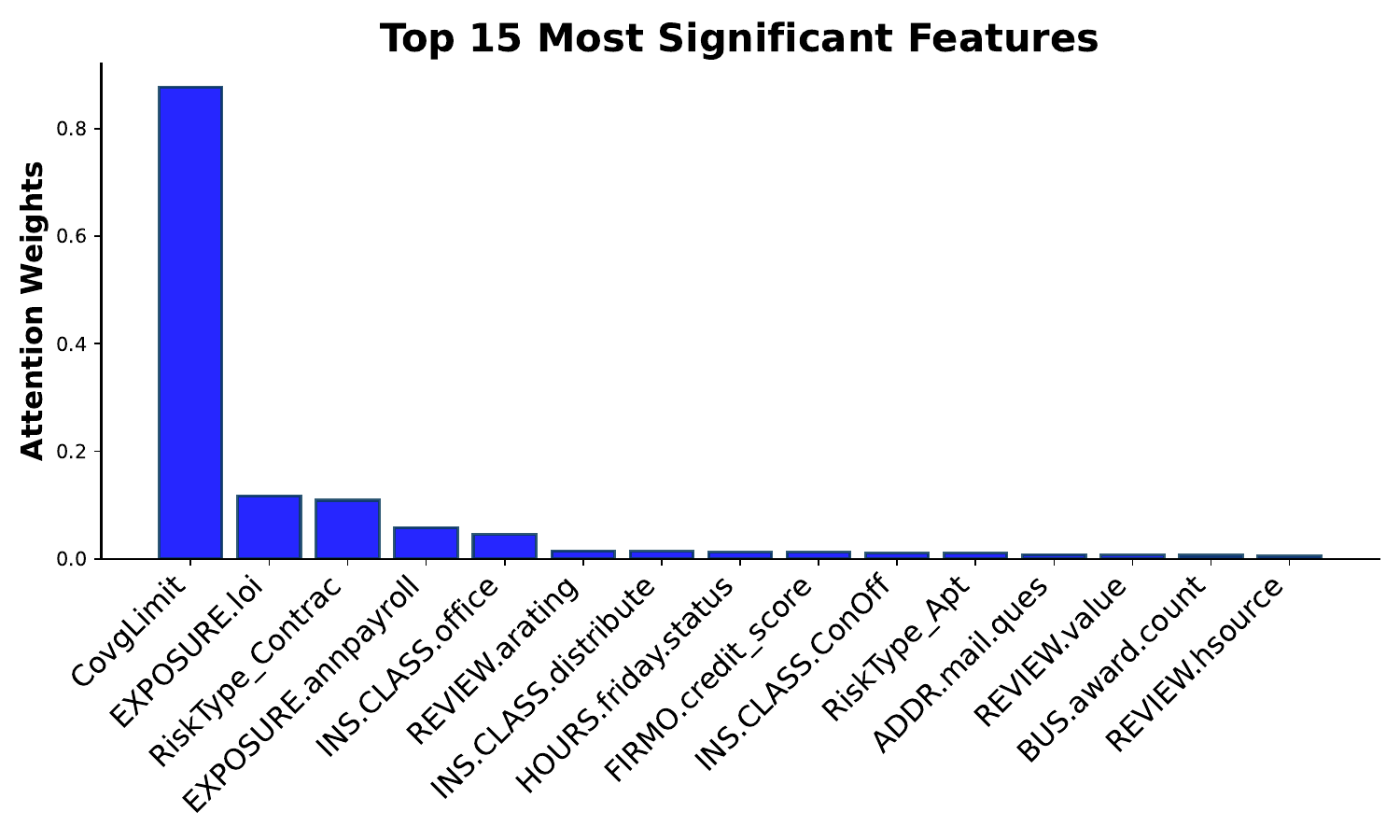}
        \caption{FCorrTransformer}
        \label{fig:top_15_feature_importance_fcorrtransformer_ins}
    \end{subfigure}
    \begin{subfigure}[t]{\textwidth}
        \centering
        \includegraphics[scale=0.4]{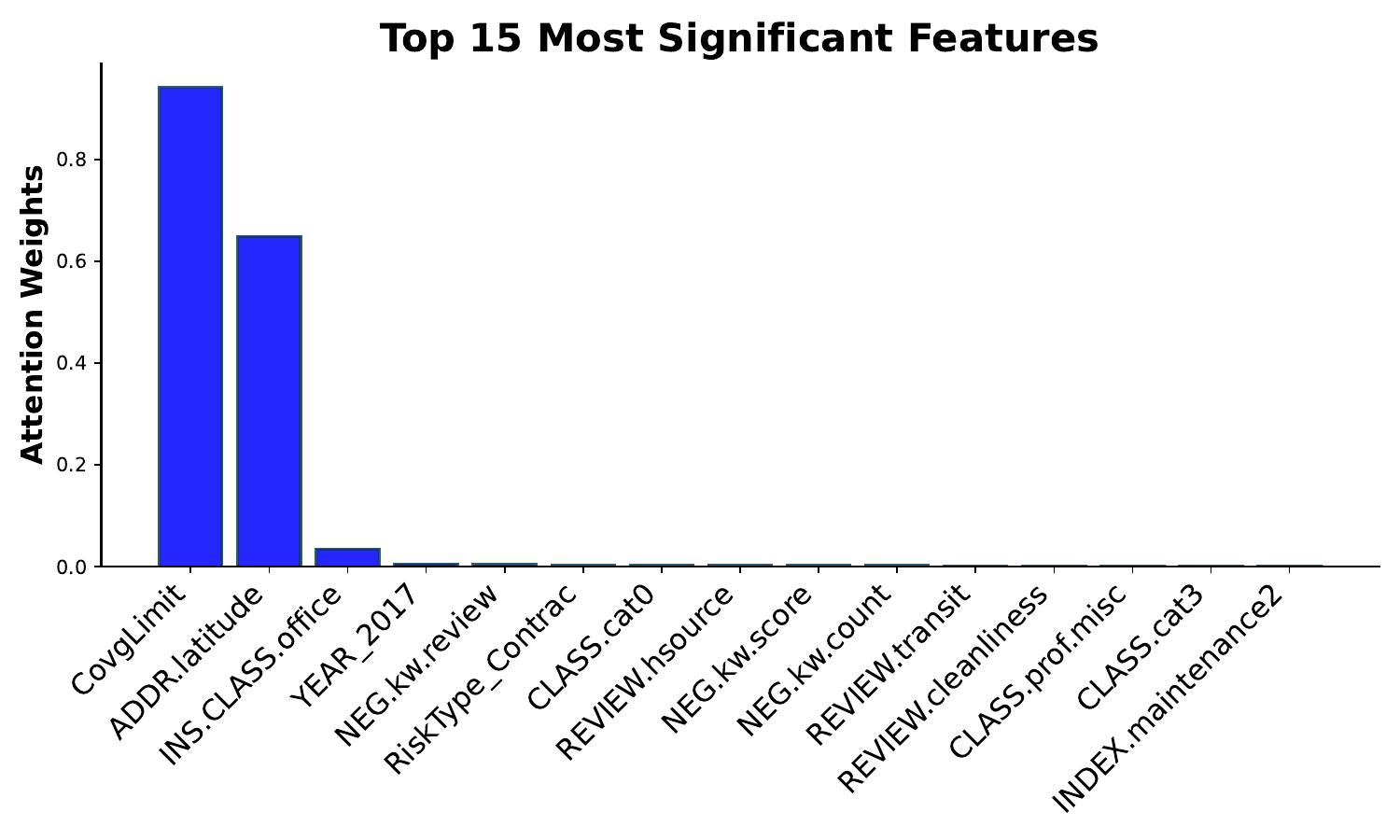}
        \caption{FCorrTransformer + CAR}
        \label{fig:top_15_feature_importance_fcorrtransformer_ins_fair}
    \end{subfigure}
    \caption{Top 15 Feature Significance on InsurTech dataset}
    \label{fig:top_15_feature_importance_InsurTech}
\end{figure}

Due to the high dimensionality of the InsurTech dataset (581 features), visualizing the full attention matrices is not informative for human interpretation. Instead, we extract the diagonal elements of the attention matrix as feature significance scores from FCorrTransformer, as shown in Figure \ref{fig:top_15_feature_importance_InsurTech}. Consistent with LightGBM, \textit{CovgLimit} (i.e., the coverage limit of the purchased insurance policy) is identified as the most influential feature, along with several features reported in \citet{quanImprovingBusinessInsurance2025}. However, the overall feature significance profiles differ across models. In the original study, territorial risk features (i.e., \textit{TERRITORY.xx}), engineered by InsurTech characterizing the density of risks at ZIP code level, are considered highly significant predictors, reflecting traditional actuarial practices that rely heavily on geographic risk segmentation. In contrast, FCorrTransformer (Figure \ref{fig:top_15_feature_importance_fcorrtransformer_ins}) assigns greater significance to insurer-provided risk characterizations (i.e., \textit{INS.CLASS.xxx}), which are strongly correlated with business credit score. These unintended attention weights are subsequently redistributed by CAR (Figure \ref{fig:top_15_feature_importance_fcorrtransformer_ins_fair}) to fairer features, indicating that CAR actively suppresses feature dependencies that may serve as proxies for sensitive socioeconomic features. As a result, attention is redistributed toward InsurTech-engineered features, such as negative keyword indicators (i.e., \textit{NEG.kw.xxx}), InsurTech classification factors (\textit{CLASS.xxx}), and engineered indices (\textit{INDEX.xxx}). These features are derived from behavioral and operational characteristics rather than legacy actuarial categorization. This shift suggests that InsurTech features, which are more personalized and fine-grained, provide relatively fairer predictive signals for insurance pricing compared to traditional firm-defined risk characteristics. More broadly, this finding highlights the potential of modern data engineering to reduce reliance on traditional risk proxies that may embed structural biases, while preserving predictive utility in regulated pricing tasks.

\begin{table}[!ht]
\centering
\caption{Complexity of Each Model on InsurTech dataset}
\label{tab:mol-complexity-InsurTech}
\begin{tabular}{cccc}
\toprule
\multirow{2}*{Model} & \multirow{2}*{Number of parameters} & \multicolumn{2}{c}{GPU Memory Usage/MB (batch of 64)} \\
 &  & Parameter size & Forward/backward pass size \\
\hline
FFN & 383,313 & 1.53 & 0.43 \\
FT-Transformer & 55,317 & 0.22 & 661.12 \\
FCorrTransformer & 21,767 & 0.09 & 2.91 \\
\bottomrule
\end{tabular}
\end{table}

The complexity analysis, summarized in Table \ref{tab:mol-complexity-InsurTech}, shows that FCorrTransformer contains roughly half the number of parameters of FT-Transformer, reflecting its attention-light design philosophy. On the InsurTech dataset, the FFN accounts for 10,145 out of 21,767 parameters (46.6\%) in FCorrTransformer, whereas it contributes less than 0.1\% of the parameters in FT-Transformer. This architectural design, combined with the high dimensionality of the InsurTech dataset, allows FCorrTransformer to use only 4.4\% of the GPU memory during training (on the same batch size), achieving more efficient training/deployment, comparable predictive performance while providing a strong counterfactual fairness guarantee.

\section{Conclusion}\label{sec:con}

In this work, we proposed a novel framework for achieving counterfactual fairness in tabular learning through the FCorrTransformer architecture and CAR. We demonstrated that an attention-light design can provide direct and interpretable representations of feature dependencies while maintaining strong predictive performance. The proposed input augmentation strategy further enables efficient counterfactual learning, making the framework scalable to high-dimensional real-world datasets. Empirical results on imbalanced financial and insurance datasets show that our framework effectively redistributes attention from biased dependencies toward fair representations. This property is particularly important in regulatory-sensitive domains such as insurance and finance, where even small discriminatory effects can result in legal liability, regulatory sanctions, or violations of consumer protection principles.

From a practical perspective, the interpretability of FCorrTransformer offers a key advantage for industry deployment. The attention matrix provides a transparent and auditable view of feature dependencies, enabling practitioners and regulators to directly inspect how sensitive features influence predictions. This transparency facilitates robust model validation, risk management, and comprehensive external audits. Moreover, the efficiency of the attention-light design makes the framework suitable for large-scale operational environments, where computational cost, latency, and model complexity are critical constraints. Beyond individual organizations, the proposed framework may also have broader implications for regulatory practice. Compared to practices such as the simple removal of sensitive features, as referenced in SB21-169, which still result in biased decision-making under counterfactual fairness evaluations, our proposed framework offers a more practical approach and provides stronger empirical support for promoting fair insurance operations. By embedding fairness directly into the model architecture and internal representations, our approach closely aligns with transparency and explainability objectives. This suggests that mechanism-based evaluations, which audit the internal logic of a model, could serve as a critical complement to existing performance-centric assessments for ML systems. 

Despite the encouraging empirical results, the proposed framework is subject to several limitations. First, in its current formulation, CAR is primarily designed for categorical sensitive features and does not naturally generalize to continuous sensitive features, such as income, for which defining counterfactual interventions remains challenging. Second, although CAR enforces empirical counterfactual invariance at the attention level without relying on explicit assumptions, it does not establish formal causal guarantees regarding the relationships between sensitive features and the prediction target. Such guarantees may be essential in settings where causal interpretability is required for regulatory or legal purposes. Finally, the attention-light architectural design of FCorrTransformer, while beneficial for interpretability and fairness control, inherently limits the expressive capacity of the model and may lead to suboptimal performance in certain nonlinear settings or tasks with extremely complex feature dependencies.

Overall, this study suggests that fairness in high-stakes domains should be embedded through explicit architectural design rather than treated as a post-hoc correction or data preprocessing step, and that attention mechanisms provide a principled and interpretable pathway toward responsible machine learning. In the context of insurance and finance, our results demonstrate the feasibility of developing predictive systems that are not only accurate, but also interpretable, and aligned with ethical principles, and broader societal expectations.

\clearpage

\section*{Acknowledgement} This research used the DeltaAI advanced computing and data resource, which is supported by the National Science Foundation (award OAC 2320345) and the State of Illinois. DeltaAI is a joint effort of the University of Illinois Urbana-Champaign and its National Center for Supercomputing Applications. This work is also supported by the Actuarial Innovation \& Technology Strategic Research (AIT) Research Grant (2025) from the Society of Actuaries (SOA).

\bibliographystyle{apalike}
\bibliography{ref.bib}

\clearpage

\begin{appendices}
\section{Notations}\label{appendix_sec:notation}

\begin{table}[!ht]
\centering
\caption{Notations used in the formulation}
\label{tab:notation}
\begin{tabular}{c c } 
\toprule
Notation & Description  \\
\midrule

$\mathcal{D}$ & a dataset \\
$(\bm{X}, \bm{y})$ & a pair of feature matrix $\bm{X}$ and response vector $\bm{y}$ \\
$N$ & number of observations \\
$\mathcal{X}$/$\mathcal{Y}$ & feature and response space \\
$\bm{X}_s$ & sensitive feature matrix \\
$\mathcal{X}_s$ & sensitive feature space \\
$\bm{X}_r$ & non-sensitive feature matrix \\
$\mathcal{X}_r$ & non-sensitive feature space \\
$p$ & number of features in feature matrix $\bm{X}$ \\
$\bm{X}_{Cat}$/$\bm{X}_{Con}$ &  categorical and continuous feature matrices \\
$p^{Cat}$/$p^{Con}$ & number of categorical and continuous features \\
$s_{i}$ & sensitive demographic category $i$ \\
$\mathcal{S}$ & sensitive demographic category space \\
$C^s$ & Number of categories for (discrete) sensitive feature \\
$f$ & a ML model \\
$C^{Cat}_i$ & Number of categories for $i$-th categorical feature \\
$\hat{\bm{y}}$ & model predictions \\
$\bm{v}$ & one-hot encoded unit vectors \\
$OH$ & one-hot encoding \\
$p_{OH}^{Cat}$ & total number of categories in categorical features \\
$GELU$ & GELU activation function \\
$ELinear$ & Element-Wise linear layer \\
$CLinear$ & CatLinear layer \\
$\bm{E}^{Cat}$/$\bm{E}^{Con}$ & categorical and continuous feature embeddings \\
$\bm{W}$/$\bm{b}$ & weight and bias parameters \\
$R(i)$ & reduction matrix \\
$CatEm$/$ConEm$ & categorical and continuous embedding layers \\
$\bm{E}$ & feature embedding \\
$LN$ & layer normalization \\
$\bm{A}$ & attention matrix \\
$\mathcal{L}_{CAR}$ & CAR loss function \\
$\sigma$ & index of sensitive feature in feature matrix \\
$\mathcal{L}$ & overall loss function \\
$\mathcal{L}_{perf}$ & performance loss function \\
$\lambda$ & regularization coefficient for CAR \\
$\mu$ & distance function for fairness evaluation \\
$\phi$ & score function for fairness evaluation \\
$\zeta$ & normalization factor for generalized fairness metrics \\
$\mathcal{D}'$ & a perturbed data \\
\bottomrule
\end{tabular}
\end{table}

\clearpage

\section{Performance Evaluation Metrics}\label{appendix_sec:performance-metric}

\begin{enumerate}
\item Accuracy Score
$$
Accuracy(\bm{y}, \hat{\bm{y}})=\dfrac{1}{N}\sum_{n=1}^{N}\mathbbm{1}_{\{y_n=\hat{y}_n\}}
$$

\item F1 Score
$$
F1(\bm{y}, \hat{\bm{y}})=2\dfrac{Pr(\bm{y}, \hat{\bm{y}})Re(\bm{y}, \hat{\bm{y}})}{Pr(\bm{y}, \hat{\bm{y}})+Re(\bm{y}, \hat{\bm{y}}}
$$
with precision score $Pr(\bm{y}, \hat{\bm{y}})=\dfrac{TP}{TP+FP}$, recall score $Re(\bm{y}, \hat{\bm{y}})=\dfrac{TP}{TP+FN}$ where true Positive $TP=\sum_{n=1}^{N}\mathbbm{1}_{\{y_{n}=1\}}\mathbbm{1}_{\{\hat{y}_{n}=1\}}$, false positive $FP=\sum_{n=1}^{N}\mathbbm{1}_{\{y_{n}=0\}}\mathbbm{1}_{\{\hat{y}_{n}=1\}}$, and false negative  $FN=\sum_{n=1}^{N}\mathbbm{1}_{\{y_{n}=1\}}\mathbbm{1}_{\{\hat{y}_{n}=0\}}$.

\item False Positive Rate (FPR)
$$
FPR(\bm{y}, \hat{\bm{y}})=\dfrac{\sum_{n=1}^{N}\mathbbm{1}_{\{y_n=0\}}\mathbbm{1}_{\{\hat{y}_n=1\}}}{\sum_{n=1}^{N}\mathbbm{1}_{\{y_n=0\}}}
$$

\item False Negative Rate (FNR)
$$
FNR(\bm{y}, \hat{\bm{y}})=\dfrac{\sum_{n=1}^{N}\mathbbm{1}_{\{y_n=1\}}\mathbbm{1}_{\{\hat{y}_n=0\}}}{\sum_{n=1}^{N}\mathbbm{1}_{\{y_n=1\}}}
$$

\item Area Under Precision-Recall Curve (AUPRC)
$$
AUPRC(\bm{y}, \hat{\bm{y}}^p)=\sum_{i}(Re^{i}(\bm{y}, \hat{\bm{y}}^p)-Re^{i-1}(\bm{y}, \hat{\bm{y}}p))Pr^{i}(\bm{y}, \hat{\bm{y}}^p)
$$
where $\hat{\bm{y}}^p$ is the predicted probabilities for the positive class, $Re_{r}^{i}$ and $Pr_{r}^{i}$ denotes the recall and precision score with $i$-th thresholds, respectively.

\item Area Under the Receiver Operating Characteristic Curve (AUROC)
$$
AUROC(\bm{y}, \hat{\bm{y}}^p)=\dfrac{\sum_{m:y_m=1}\sum_{n:y_n=0}\mathbbm{1}_{\{\hat{y}_m^p>\hat{y}_n^p\}}}{\sum_{n=1}^{N}\mathbbm{1}_{\{y_{n}=1\}}\sum_{n=1}^{N}\mathbbm{1}_{\{y_{n}=0\}}}
$$

\item Gini index 
$$
Gini(\bm{y}, \hat{\bm{y}})= 1 - \dfrac{2}{N-1} \left(N - \dfrac{\sum_{n=1}^N \; ny_{[n]}}{\sum_{n=1}^N ny_{[n]}}\right)
$$
where $y_{[n]}$ is $n$-th value of $\bm{y}$ based on ranks of $\hat{\bm{y}}$.

\item Percentage Error (PE)
$$
PE(\bm{y}, \hat{\bm{y}})=\dfrac{\sum_{n=1}^N(\hat{y}_n-y_n)}{\sum_{n=1}^Ny_n}
$$

\item Root Mean Squared Error (RMSE)
$$
RMSE(\bm{y}, \hat{\bm{y}})=\sqrt{\dfrac{1}{N}\sum_{n=1}^N(\hat{y}_n-y_n)^2}
$$

\item Mean Absolute Error (MAE)
$$
MAE(\bm{y}, \hat{\bm{y}})=\dfrac{1}{N}\sum_{n=1}^N|\hat{y}_n-y_n|
$$

\end{enumerate}

\clearpage

\section{Fairness Evaluation Metrics}\label{appendix_sec:performance-fair}

\begin{enumerate}
\item Demographic Parity Difference (DPD) \citep{agarwalReductionsApproachFair2018}
$$
DPD(f, \mathcal{D})=\underset{s_{j},s_{k}}{\max}|\mathbb{E}(\hat{\bm{y}}|\bm{X}_s=s_j)-\mathbb{E}(\hat{\bm{y}}|\bm{X}_s=s_k)|
$$

\item Equalized Odds Difference (EqOdd) \citep{hardtEqualityOpportunitySupervised2016}
$$
EqOdd(f, \mathcal{D})=\underset{s_{j},s_{k}}{\max}\underset{y\in\{0, 1\}}{\max}|P(\hat{\bm{y}}=1|\bm{X}_s=s_{j}, \bm{y}=y)-P(\hat{\bm{y}}=1|\bm{X}_s=s_{k}, \bm{y}=y)|
$$

\item Equality of Opportunity Difference (EqOpp) \citep{hardtEqualityOpportunitySupervised2016}
$$
EqOpp(f, \mathcal{D})=\underset{s_{j},s_{k}}{\max}|P(\hat{\bm{y}}=1|\bm{X}_s=s_{j}, \bm{y}=1)-P(\hat{\bm{y}}=1|\bm{X}_s=s_{k}, \bm{y}=1)|
$$

\item Average Individual Fairness (AvgIF)
\citep{huangReducingSentimentBias2020} 
$$
AvgIF(f, \mathcal{D})=\frac{1}{C^s (C^s-1)}\sum_{s_{i}\in\mathcal{S}}\sum_{\substack{s_{j},s_{k}\in\mathcal{S}\\ s_{j}\neq s_{k}}}\mathcal{W}_1(\hat{\bm{y}}'_{s_{i}\rightarrow s_{j}}, \hat{\bm{y}}'_{s_{i}\rightarrow s_{k}})
$$
where $\hat{\bm{y}}'_{s_{i}\rightarrow s_{j}}=f(\mathcal{D}'_{s_{i}\rightarrow s_{j}})$ and $\hat{\bm{y}}'_{s_{i}\rightarrow s_{k}}=f(\mathcal{D}'_{s_{i}\rightarrow s_{k}})$ are the predictions of the corresponding subsets in the perturbed dataset, and $\mathcal{W}_1$ denotes the Wasserstein-1 distance measuring the difference between two distributions.

\item F1 Gap
$$
F1\_Gap(f, \mathcal{D})=\frac{1}{C^s (C^s-1)}\sum_{s_{i}\in\mathcal{S}}\sum_{\substack{s_{j},s_{k}\in\mathcal{S}\\ s_{j}\neq s_{k}}}|F1(\bm{y}'_{s_{i}\rightarrow s_{j}}, \hat{\bm{y}}'_{s_{i}\rightarrow s_{j}})-F1(\bm{y}'_{s_{i}\rightarrow s_{k}}, \hat{\bm{y}}'_{s_{i}\rightarrow s_{k}})|
$$

\item AUROC Gap
{\footnotesize
$$
AUROC\_Gap(f, \mathcal{D})=\frac{1}{C^s (C^s-1)}\sum_{s_{i}\in\mathcal{S}}\sum_{\substack{s_{j},s_{k}\in\mathcal{S}\\ s_{j}\neq s_{k}}}|AUROC(\bm{y}'_{s_{i}\rightarrow s_{j}}, \hat{\bm{y}}'_{s_{i}\rightarrow s_{j}})-AUROC(\bm{y}'_{s_{i}\rightarrow s_{k}}, \hat{\bm{y}}'_{s_{i}\rightarrow s_{k}})|
$$
}

\item AUPRC Gap
{\footnotesize
$$
AUPRC\_Gap(f, \mathcal{D})=\frac{1}{C^s (C^s-1)}\sum_{s_{i}\in\mathcal{S}}\sum_{\substack{s_{j},s_{k}\in\mathcal{S}\\ s_{j}\neq s_{k}}}|AUPRC(\bm{y}'_{s_{i}\rightarrow s_{j}}, \hat{\bm{y}}'_{s_{i}\rightarrow s_{j}})-AUPRC(\bm{y}'_{s_{i}\rightarrow s_{k}}, \hat{\bm{y}}'_{s_{i}\rightarrow s_{k}})|
$$
}

\item RMSE Gap
{\footnotesize
$$
RMSE\_Gap(f, \mathcal{D})=\frac{1}{C^s (C^s-1)}\sum_{s_{i}\in\mathcal{S}}\sum_{\substack{s_{j},s_{k}\in\mathcal{S}\\ s_{j}\neq s_{k}}}|RMSE(\bm{y}'_{s_{i}\rightarrow s_{j}}, \hat{\bm{y}}'_{s_{i}\rightarrow s_{j}})-RMSE(\bm{y}'_{s_{i}\rightarrow s_{k}}, \hat{\bm{y}}'_{s_{i}\rightarrow s_{k}})|
$$
}

\item MAE Gap
{\footnotesize
$$
MAE\_Gap(f, \mathcal{D})=\frac{1}{C^s (C^s-1)}\sum_{s_{i}\in\mathcal{S}}\sum_{\substack{s_{j},s_{k}\in\mathcal{S}\\ s_{j}\neq s_{k}}}|MAE(\bm{y}'_{s_{i}\rightarrow s_{j}}, \hat{\bm{y}}'_{s_{i}\rightarrow s_{j}})-MAE(\bm{y}'_{s_{i}\rightarrow s_{k}}, \hat{\bm{y}}'_{s_{i}\rightarrow s_{k}})|
$$
}

\end{enumerate}

\clearpage

\section{Details on Experiments}\label{appendix_sec:detail}

\subsection{Imbalanced Classification: BAF}\label{appendix_subsec:bak}

In the BAF experiment, to establish an independent decision threshold suitable for imbalanced classification—since the default threshold of 0.5 is not appropriate—and to provide a strong performance baseline, we train a LightGBM model \citep{keLightGBMHighlyEfficient2017} with default hyperparameter settings. The decision threshold is selected by maximizing the training F1 score of the LightGBM model, resulting in a threshold of 0.12. This partially explains the superior performance of LightGBM in terms of Accuracy, FPR, and FNR, as these metrics are highly sensitive to the choice of decision threshold in imbalanced classification settings. The performance of the LightGBM baseline is summarized in Table \ref{tab:performance-lightgbm}, and it is regarded as one of the most stable and best-performing models in tabular learning.

\begin{table}[!ht]
\centering
\caption{More Model Baseline Comparison}
\label{tab:performance-lightgbm}
\footnotesize
\begin{tabular}{cccccccc}
\toprule
Subset & Model & Accuracy & F1 score & FPR & FNR & AUROC & AUPRC \\
\hline
\multirow{3}*{Train} & LightGBM & \textbf{0.9832} & \textbf{0.3380} & \textbf{0.0101 }& \textbf{0.6149} & \textbf{0.9273} & \textbf{0.2644} \\
& FCorrTransformer + Res & 0.9807 & 0.2423 & 0.0114 & 0.7229 & 0.8966 & 0.1743 \\
& FCorrTransformer + Rem & 0.9771 & 0.2272 & 0.0153 & 0.6976 & 0.8860 & 0.1527 \\
& FCorrTransformer & 0.9824 & 0.2343 & 0.0093 & 0.7578 & 0.8928 & 0.1648 \\
\hline
\multirow{3}*{Test} & LightGBM & \textbf{0.9812} & 0.2267 & \textbf{0.0112} & 0.7383 & 0.8953 & 0.1442 \\
& FCorrTransformer + Res & 0.9809 & \textbf{0.2326} & 0.0115 & \textbf{0.7259} & \textbf{0.8961} & \textbf{0.1526} \\
& FCorrTransformer + Rem & 0.9774 & 0.2202 & 0.0154 & 0.6970 & 0.8826 & 0.1478\\
& FCorrTransformer & 0.9826 & 0.2220 & 0.0094 & 0.7648 & 0.8925 & 0.1458 \\
\bottomrule
\end{tabular}
\end{table}

\begin{figure}[!ht]
    \centering
    \includegraphics[scale = 0.44]{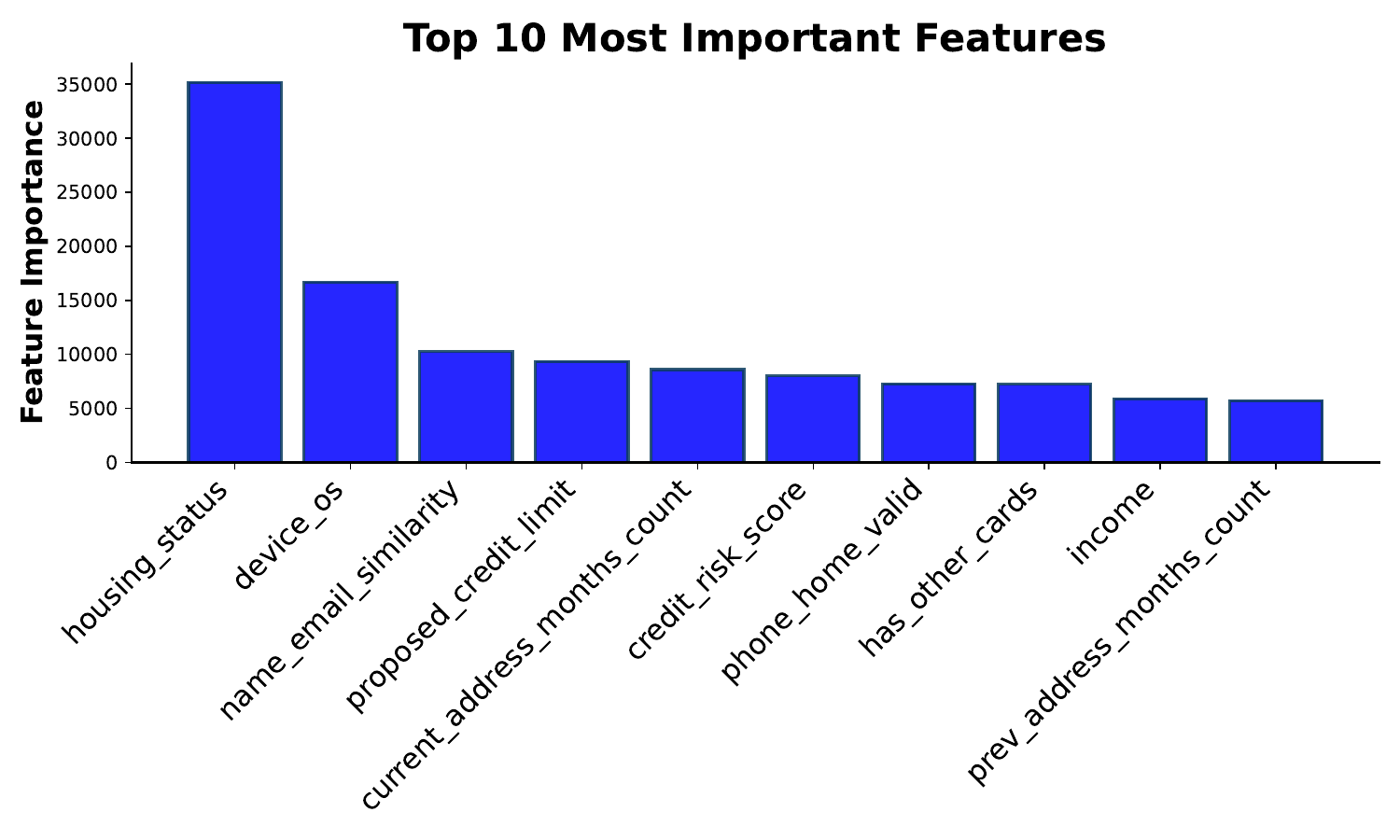}
    \caption{Top 10 Feature Significance in LightGBM on BAF dataset}
    \label{fig:top_10_feature_importance_lightgbm}
\end{figure}

To assess whether the performance gap of FCorrTransformer is caused by its simplified architecture, we reintroduce the residual connection in the attention mechanism (the dashed connection in Figure \ref{fig:FCorr_architecture}), and denote this variant as FCorrTransformer + Res in Table \ref{tab:performance-lightgbm}. The results show that adding the residual connection significantly improves performance, making it comparable to LightGBM on the test set. However, this modification largely negates the effect of CAR and weakens the counterfactual fairness guarantees, as the residual path allows biased information to bypass the attention redistribution mechanism. Furthermore, motivated by the formulation of counterfactual fairness metrics and the structure of causal models used to capture the influence of sensitive features, a straightforward strategy is to simply remove the sensitive feature and enforce the model to be invariant to it. However, such naive feature removal often leads to noticeable performance drop (as illustrated in Table \ref{tab:performance-lightgbm} for FCorrTransformer + Rem) or fails to eliminate bias due to indirect discrimination, where correlated non-sensitive features still encode sensitive information. In our experiments, this approach results in a DPD of 0.0769 and EqOdd/EqOpp of 0.6516 on the training set, which are significantly higher than those achieved by FCorrTransformer + CAR in Table \ref{tab:fairness-BAF}. For comparison with the proposed framework, we visualize the top-10 feature importance scores obtained from LightGBM’s built-in feature importance function in Figure \ref{fig:top_10_feature_importance_lightgbm}. Among the 31 features, five features (in the top-10 list) overlap with those identified by FCorrTransformer in Figures \ref{fig:top_10_feature_importance_fcorrtransformer} and \ref{fig:top_10_feature_importance_fcorrtransformer_fair}. This suggests that although FCorrTransformer learns a slightly different internal representation from LightGBM, the two models are largely consistent in identifying the most informative features. One thing to note about the performance of TabTransformer in Table \ref{tab:performance-BAF} is that, due to its architectural design, the model size cannot be reduced to be comparable with the other models. Therefore, we adopt a minimal TabTransformer configuration for this dataset. However, despite its relatively large number of parameters, TabTransformer does not achieve better performance in our experiments.

\subsection{Classification: Adult}\label{appendix_subsec:adult}

We further evaluate the proposed framework on the well-studied Adult (Census Income) dataset from the UCI Machine Learning Repository\footnote{\url{https://archive.ics.uci.edu/dataset/20/census+income}}. The model’s predictive performance and fairness metrics are summarized in Tables \ref{tab:performance-adult} and \ref{tab:fairness-adult}, respectively. The results indicate that FCorrTransformer achieves predictive performance comparable to FT-Transformer, while simultaneously improving counterfactual fairness through the CAR regularization. These findings further strengthen the feasibility and practical applicability of the proposed framework.

\begin{table}[!ht]
\centering
\caption{Model Performance Comparison on Adult dataset}
\label{tab:performance-adult}
\footnotesize
\begin{tabular}{cccccccc}
\toprule
Subset & Model & Accuracy & F1 score & FPR & FNR & AUROC & AUPRC \\
\hline
\multirow{4}*{Train} & FT-Transformer & 0.8570 & 0.6628 & \textbf{0.0562} & 0.4165 & 0.9158 & 0.7753 \\
& FT-Transformer + CAR & 0.8602 & \textbf{0.6935} & 0.0753 & 0.3432 & 0.9163 & 0.7742 \\
& FCorrTransformer & \textbf{0.8600} & 0.6894 & 0.0718 & 0.3549 & \textbf{0.9178} & \textbf{0.7973} \\
& FCorrTransformer + CAR &  0.8388 & 0.6806 & 0.1214 & \textbf{0.2867} & 0.9029 & 0.7461 \\
\hline
\multirow{4}*{Test} & FT-Transformer & 0.8476 & 0.6346 & \textbf{0.0636} & 0.4397 & 0.9011 & 0.7439 \\
& FT-Transformer + CAR & 0.8466 & 0.6565 & 0.0836 & 0.3794 & 0.8995 & 0.7372 \\
& FCorrTransformer & \textbf{0.8573} & \textbf{0.6781} & 0.0743 & 0.3638 & \textbf{0.9128} & \textbf{0.7835} \\
& FCorrTransformer + CAR & 0.8369 & 0.6704 & 0.1215 & \textbf{0.2977} & 0.8994 & 0.7344 \\
\bottomrule
\end{tabular}
\end{table}

\begin{table}[!ht]
\centering
\caption{Fairness Metrics Comparison on Adult dataset}
\label{tab:fairness-adult}
\scriptsize
\begin{tabular}{ccccccccc}
\toprule
Subset & Model & DPD & EqOdd & EqOpp & AvgIF & F1 Gap & AUROC Gap & AUPRC Gap \\
\hline
\multirow{6}*{Train} & FT-Transformer & 0.1850 & 0.3200 & 0.3200 & 0.1303 & 0.2474 & 0.0723 & 0.2542 \\
& FT-Transformer + CAR & 0.1886 & 0.2838 & 0.2838 & 0.1165 & 0.3083 & 0.0489 & 0.2112 \\
& FCorrTransformer & \textbf{0.1705} & \textbf{0.1818} & \textbf{0.1818} & 0.0994 & 0.1353 & 0.0097 & 0.0254 \\
& FCorrTransformer + CAR & 0.1851 & 0.2246 & 0.2246 & \textbf{0.0001} & \textbf{0.0000} & \textbf{0.0000} & \textbf{0.0000} \\
\hline
\multirow{6}*{Test} & FT-Transformer & 0.1914 & 0.4286 & 0.4286 & 0.1362 & 0.4086 & 0.0794 & 0.2902 \\
& FT-Transformer + CAR & 0.2017 & 0.3910 & 0.3910 & 0.1150 & 0.4174 & 0.0448 & 0.1554 \\
& FCorrTransformer & \textbf{0.1766} &  0.2180 & 0.2180 & 0.1013 & 0.1921 & 0.0087 & 0.0286 \\
& FCorrTransformer + CAR & 0.1901 & \textbf{0.1694} & \textbf{0.1694} & \textbf{0.0001} & \textbf{0.0000} & \textbf{0.0000} & \textbf{0.0004} \\
\bottomrule
\end{tabular}
\end{table}

\clearpage

\section{Controllability of Counterfactual Fairness}\label{appendix_sec:controllability}

In practice, rather than enforcing uniformly strong counterfactual fairness requirements across all insurance workflows, regulators may impose different levels of fairness requirements depending on the specific business function. For instance, stricter counterfactual fairness standards may be required for insurance pricing, while comparatively less stringent requirements may apply to tasks such as customer retention or risk management. Consequently, the ability to control the degree of fairness can be practically valuable in real-world applications. Given the regularization-based formulation of CAR, a natural approach is to adjust the regularization coefficient $\lambda$ to regulate the level of counterfactual fairness imposed during model training.

\begin{figure}[!ht]
    \centering
    \includegraphics[scale = 0.4]{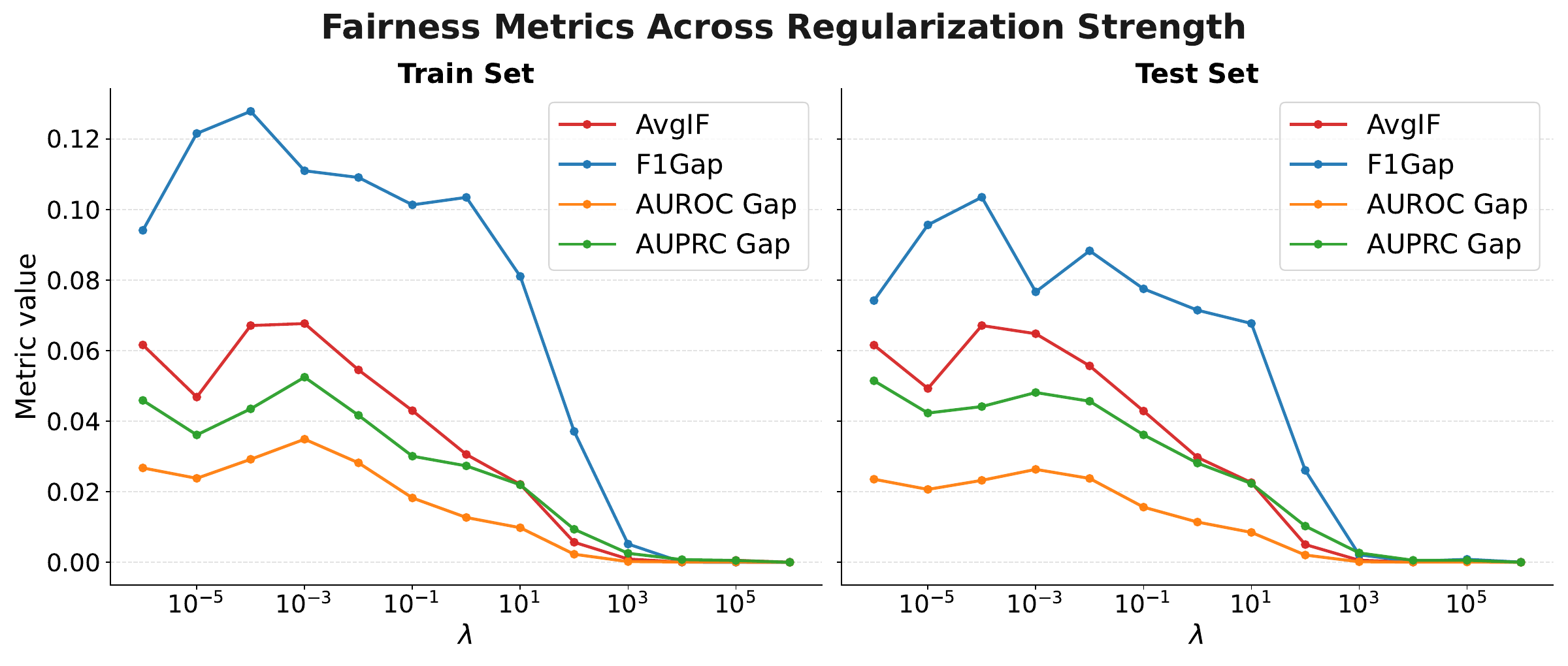}
    \caption{Controllability of Counterfactual Fairness through CAR}
    \label{fig:controllability_fairness}
\end{figure}

Figure \ref{fig:controllability_fairness} illustrates the behavior of counterfactual fairness metrics under varying values of the regularization coefficient for the imbalanced classification task introduced in Subsection \ref{subsec:bak}. Overall, the results align with expectations: increasing the regularization strength imposes stronger constraints on the attention matrices, leading to improved counterfactual fairness outcomes. When the regularization coefficient $\lambda$ is small, its impact on the fairness metrics remains limited. As $\lambda$ increases, the effect becomes more substantial. In particular, the AvgIF metric and the AUROC/AUPRC gap, which capture distributional differences across sensitive demographic categories, exhibit relatively stable and monotonic controllability as regularization increases. In contrast, the F1 gap shows substantially greater variability and weaker controllability. This behavior can be attributed to the sensitivity of F1-based metrics to decision threshold selection, especially in highly imbalanced learning settings. These observations suggest that CAR generally enables practitioners to adjust counterfactual fairness through the regularization coefficient to accommodate varying fairness requirements across applications. However, careful consideration is necessary when fairness metrics are strongly influenced by model calibration or threshold-dependent evaluation procedures.

\clearpage

\section{Discussion on Multi Encoder Layer FCorrTransformer}\label{appendix_sec:multi-layer}

Under the formulation of the dot-product attention mechanism, the elements of the first-layer attention matrix represent pairwise attention scores between features. In deeper layers, however, the attention matrices become a mixture of previous attention scores based on all features, making them more difficult to interpret as the original feature dependency structure gradually degenerates. Furthermore, applying fairness regularization directly to the full attention matrices often leads to noticeable performance degradation, as observed in our preliminary experiments. In contrast, the design of FCorrTransformer imposes sparsity on the attention matrices through low-dimensional embedding, allowing deeper-layer attention elements to remain largely dominated by the corresponding pairwise feature dependencies. This property helps preserve the interpretability of feature relationships compared with the dense attention matrices typically observed in architectures such as FT-Transformer. For the regularization, the CAR term in Equation \ref{eq:car-cia} can be modified to operate on the final-layer attention matrix. However, empirical experiments indicate that counterfactual fairness regularized with the final attention layer is difficult to achieve in FCorrTransformer architectures that have more than two encoder layers. This behavior is expected, as the composition of attention matrices in deeper layers becomes increasingly complex and therefore harder to regulate through CAR. Consequently, for multi-layer configurations, we retain the first-layer regularization defined in Equation \ref{eq:car-cia}, where the regularization directly operates on pairwise feature dependencies.

\begin{figure}[!ht]
    \centering
    \begin{subfigure}[b]{0.48\textwidth}
        \centering
        \includegraphics[width=\textwidth]{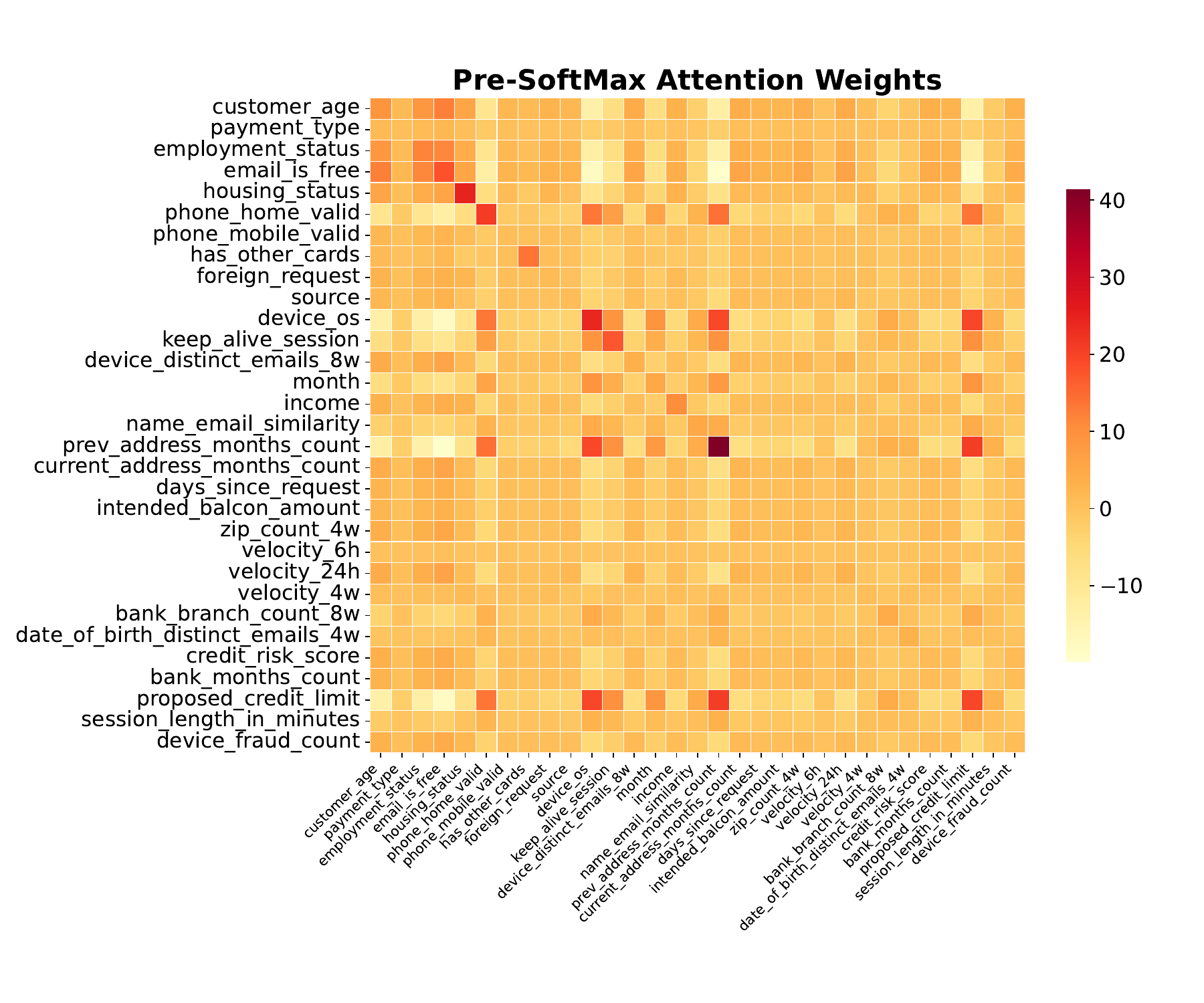}
        \caption{First-layer attention in 3-layer FCorrTransformer}
        \label{fig:attn_3_1}
    \end{subfigure}
    \begin{subfigure}[b]{0.48\textwidth}
        \centering
        \includegraphics[width=\textwidth]{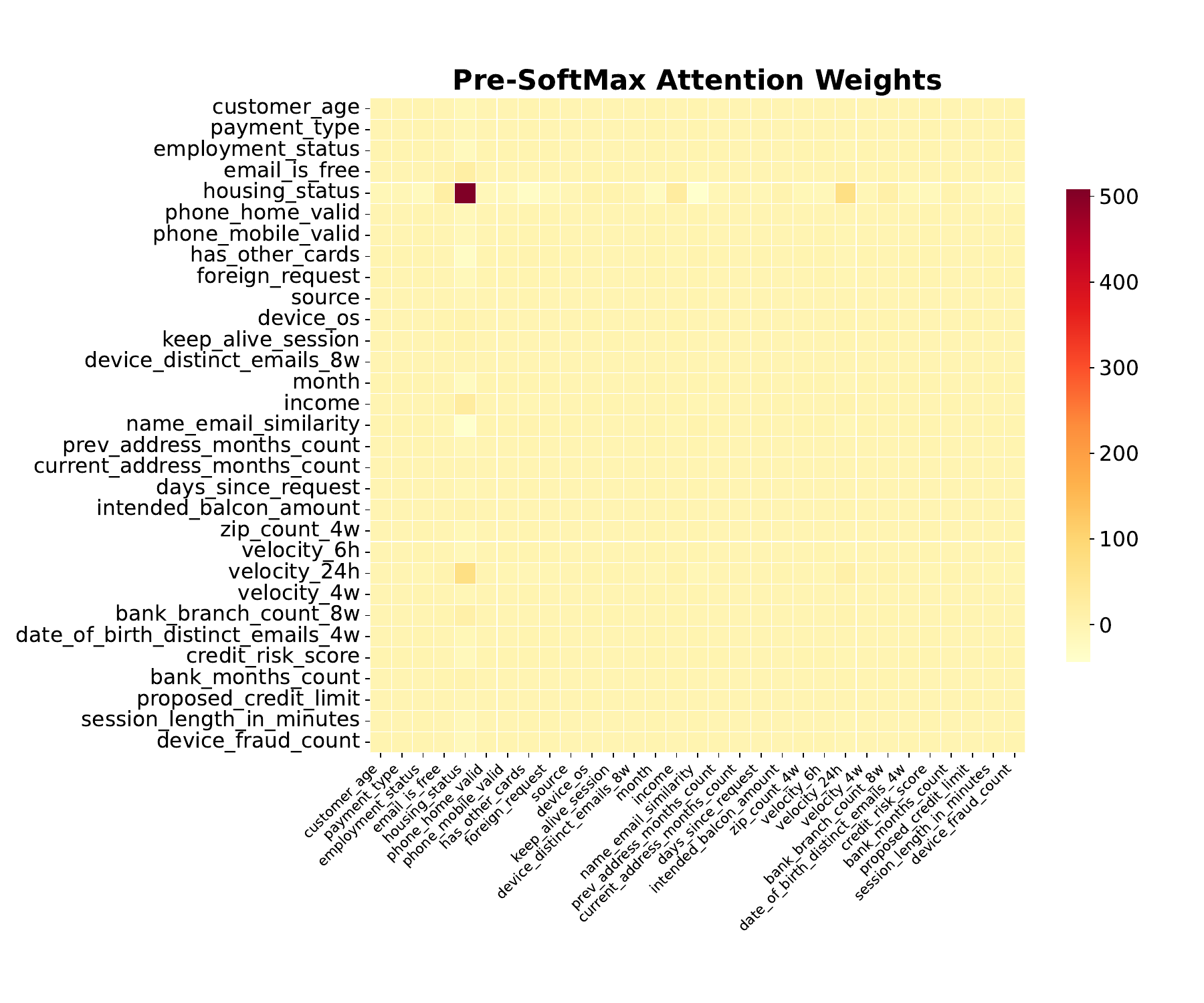}
        \caption{First-layer attention in 5-layer FCorrTransformer}
        \label{fig:attn_5_1}
    \end{subfigure}
    
    \begin{subfigure}[b]{0.48\textwidth}
        \centering
        \includegraphics[width=\textwidth]{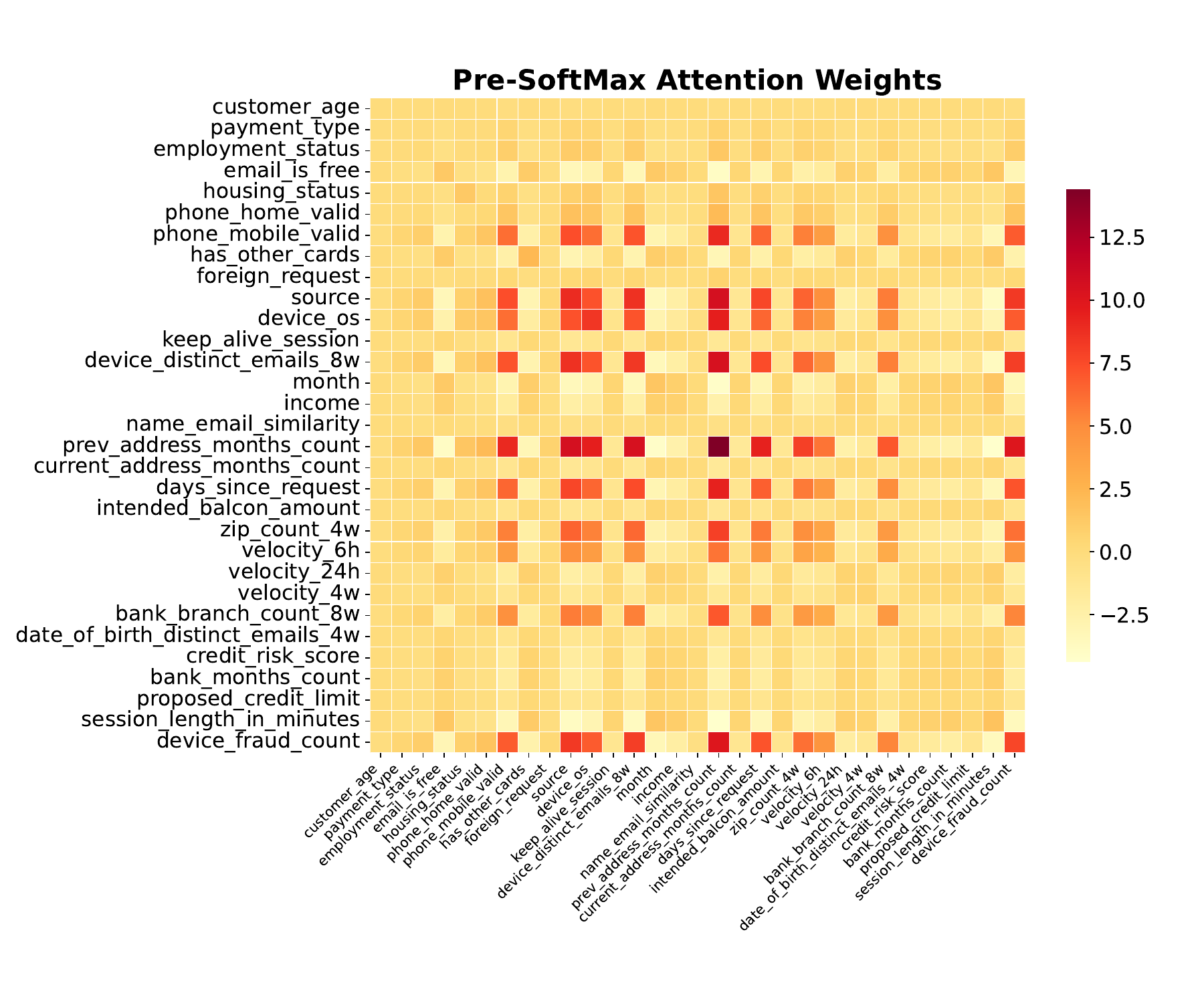}
        \caption{Third-layer attention in 3-layer FCorrTransformer}
        \label{fig:attn_3_3}
    \end{subfigure}
    \begin{subfigure}[b]{0.48\textwidth}
        \centering
        \includegraphics[width=\textwidth]{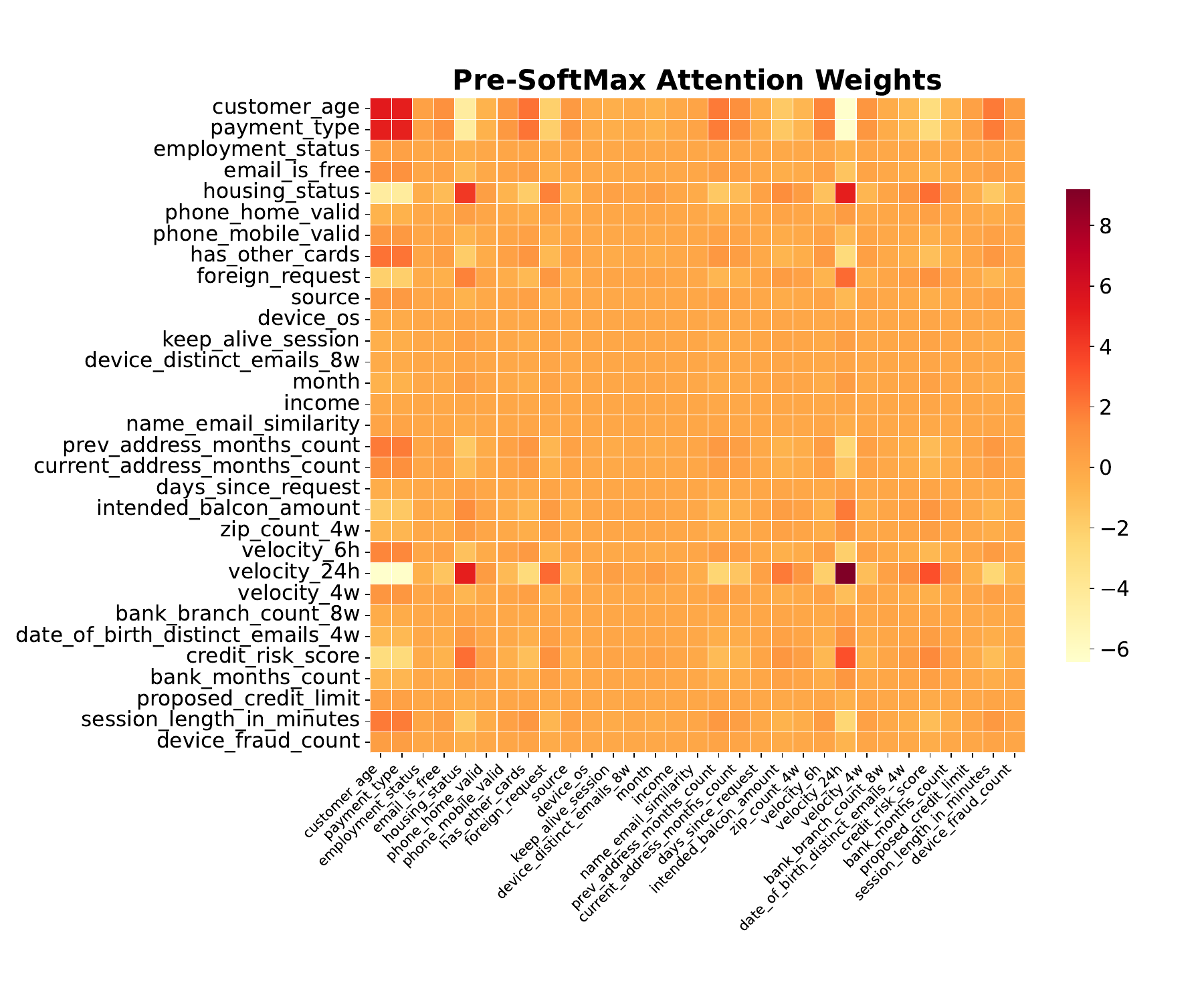}
        \caption{Fifth-layer attention in 5-layer FCorrTransformer}
        \label{fig:attn_5_5}
    \end{subfigure}
    
    \caption{Attention Heatmaps for the Multi-layer FCorrTransformer + CAR}
    \label{fig:multi-layer-fcorr}
\end{figure}

As an illustrative example, we construct 3-layer and 5-layer FCorrTransformer models trained on the BAF dataset described in Subsection \ref{subsec:bak}, using identical hyperparameters except for the number of encoder layers. We omit the predictive performance metrics, as the multi-layer models consistently underperform in our experiments. Table \ref{tab:multi-layer} summarizes the fairness metrics for the models with different encoder depths, while Figure \ref{fig:multi-layer-fcorr} visualizes the first- and last-layer attention matrices of the 3-layer and 5-layer architectures. The results indicate that CAR can still achieve counterfactual fairness under multi-layer settings. However, this improvement generally comes at the expense of both predictive performance and interpretability. In particular, attention matrices in deeper layers cannot be interpreted as pairwise feature dependencies. For this reason, we recommend adopting a single-layer FCorrTransformer architecture in practice.
Furthermore, in our empirical experiments, FCorrTransformer generally favors shallow model architectures. We attribute this behavior to the architectural design, in which the raw feature values remain critical for projecting inputs to the response space. Accordingly, FCorrTransformer employs a lightweight attention mechanism to preprocess the input features and capture their dependencies, while the feedforward layers perform the primary transformations required to map features to the response. This design differs from many existing tabular transformer architectures, where features are first projected into high-dimensional embedding spaces to capture complex dependencies, and the subsequent feedforward layers play a relatively lighter role in the prediction process.

\begin{table}[!ht]
\centering
\caption{Fairness Metrics with Multi-layer FcorrTransformer + CAR on BAF dataset}
\label{tab:multi-layer}
\scriptsize
\begin{tabular}{ccccccccc}
\toprule
Subset & \# Encoder & DPD & EqOdd & EqOpp & AvgIF & F1 Gap & AUROC Gap & AUPRC Gap \\
\hline
\multirow{3}*{Train} & 1 & 0.0456 & 0.2325 & 0.2325 & 0.0000 & 0.0000 & 0.0000 & 0.0000 \\
& 3 & 0.0736 & 0.2839 & 0.2839 & 0.0000 & 0.0000 & 0.0000 & 0.0000 \\
& 5 & 0.1005 & 0.2033 & 0.2033 & 0.0000 & 0.0000 & 0.0000 & 0.0000 \\
\hline
\multirow{3}*{Test} & 1 & 0.0859 & 0.7143 & 0.7143 & 0.0000 & 0.0000 & 0.0000 & 0.0000 \\
& 3 & 0.1794 & 0.7143 & 0.7143 & 0.0000 & 0.0000 & 0.0000 & 0.0000 \\
& 5 & 0.0625 & 0.5714 & 0.5714 & 0.0000 & 0.0000 & 0.0000 & 0.0000 \\
\bottomrule
\end{tabular}
\end{table}

\end{appendices}

\end{document}